\documentclass[journal,comsoc,xcdraw,table]{IEEEtran}
\usepackage[T1]{fontenc}

\ifCLASSINFOpdf
\else
\fi
\usepackage{amsmath}
\usepackage[cmintegrals]{newtxmath}
\usepackage{tikz}
\usepackage{amsmath}
\usepackage{booktabs}
\usepackage{multirow}
\usepackage{graphicx}
\usepackage{enumitem}
\usepackage{amsfonts}
\usepackage{tcolorbox}
\usepackage{url}
\tcbuselibrary{breakable}
\usepackage[subtle]{savetrees}

\begin{document}
%
\title{IPatch: A Remote Adversarial Patch}
%
%
%

\author{Yisroel~Mirsky
\thanks{The author is with Ben-Gurion University (e-mail: yisroel@post.bgu.ac.il see https://ymirsky.github.io/).}
}

%
%

\markboth{Journal of \LaTeX\ Class Files,~Vol.~XX, No.~XX, Month~XXXX}%
{Yisroel Mirsky: IPatch: A Remote Adversarial Patch}
%



\maketitle

\begin{abstract}
Applications such as autonomous vehicles and medical screening use deep learning models to localize and identify hundreds of objects in a single frame. In the past, it has been shown how an attacker can fool these models by placing an adversarial patch within a scene. However, these patches must be placed in the target location and do not explicitly alter the semantics elsewhere in the image.

In this paper, we introduce a new type of adversarial patch which alters a model's perception of an image's semantics. These patches can be placed anywhere within an image to change the classification or semantics of locations far from the patch. We call this new class of adversarial examples `remote adversarial patches' (RAP).

We implement our own RAP called IPatch and perform an in-depth analysis on image segmentation RAP attacks using five state-of-the-art architectures with eight different encoders on the CamVid street view dataset. Moreover, we demonstrate that the attack can be extended to object recognition models with preliminary results on the popular YOLOv3 model.  We found that the patch can change the classification of a remote target region with a success rate of up to 93\% on average. 
\end{abstract}


%
\IEEEpeerreviewmaketitle

\section{Introduction}


Deep learning has become the go-to method for automating image-based tasks.
This is because, deep neural networks (DNNs) are excellent at learning and identifying spatial patterns and abstract concepts. With advances in both hardware and neural architectures, deep learning has become both a practical and reliable solution. Companies now use image-based deep learning to automate tasks in life critical operations such as autonomous driving \cite{WaymoRem80:online,Autopilo49:online}, surveillance \cite{Artifici74:online}, and medical image screening \cite{DeepReso39:online}. 

In tasks such as these, multiple objects must be identified per image. One way to accomplish this is to predict a class probability for each pixel in the input image $x$. This approach is called image segmentation and companies such as Telsa use it to guide their autonomous vehicles safely through an environment \cite{Autopilo49:online}. Another approach is called object detection where $x$ is split into a grid of cells or regions and the model predicts both a class probability and a bounding box for each of them \cite{redmon2018yolov3,ren2015faster}. In both cases, these models rely on image semantics to successfully parse and interpret a scene.

\begin{figure}[t]
	\centering
	\includegraphics[width=\columnwidth]{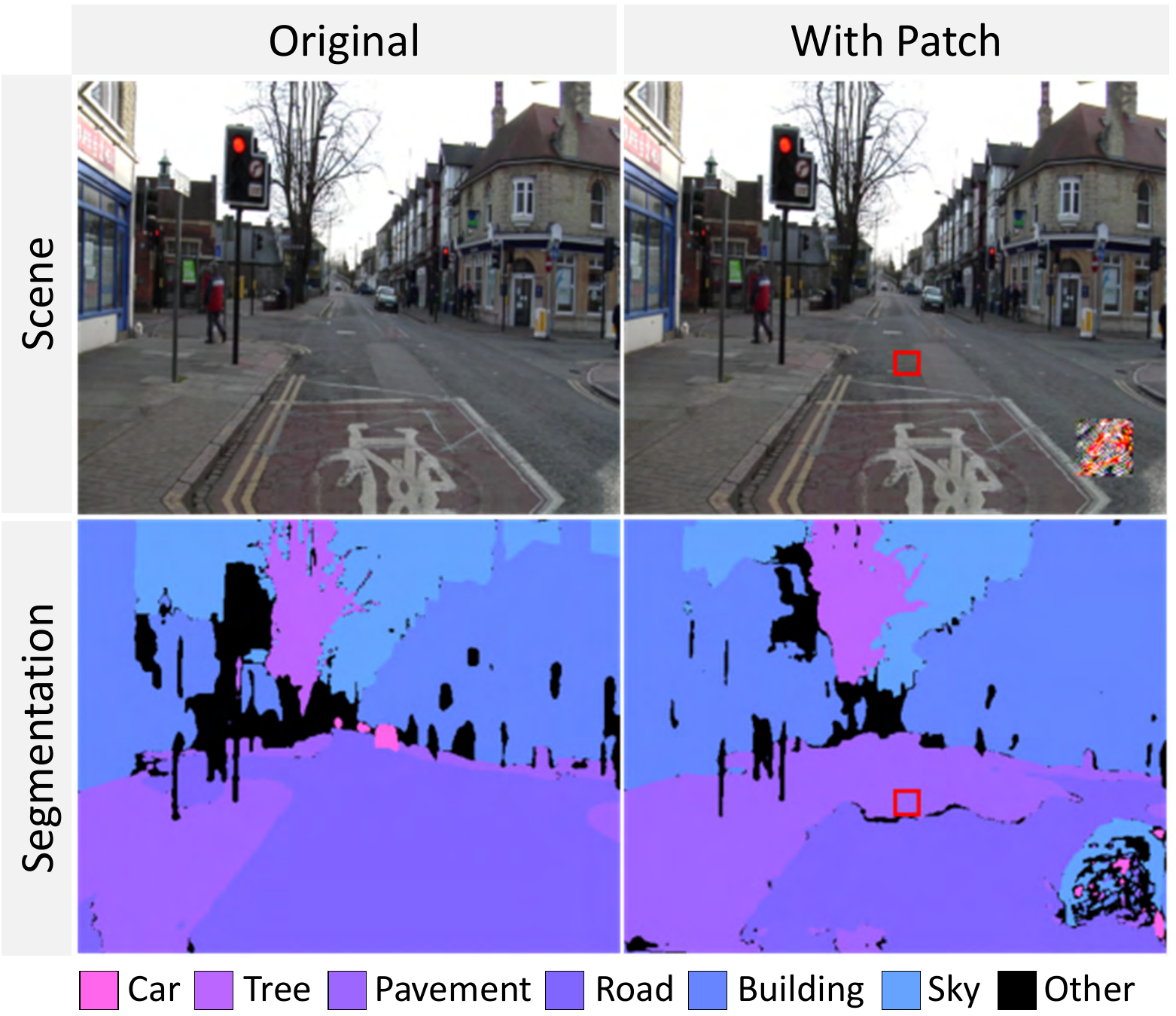}
	\caption{An example of a remote adversarial attack on a segmentation model. Here the patch has been designed to alter a predetermined location (the red box) to the class of `pavement'. The same patch works on different images and with different positions and scales.}
	\label{fig:targeted}
\end{figure}

Just like other deep learning models, these semantic models are also susceptible to adversarial attacks. In 2017, researchers demonstrated how a small `adversarial' patch can be placed in a real world scene and override an image-classifier's prediction, regardless of the patch's location or orientation \cite{brown2017adversarial}. This gave rise to a number of works which demonstrated the concept of adversarial patches against image segmentation and object detection models \cite{song2018physical,liu2018dpatch,chen2018shapeshifter,sitawarin2018darts,lee2019physical,thys2019fooling,zhao2019seeing,li2020adaptive,den2020adversarial,huang2020universal,hoory2020dynamic,wu2020making}. However, current adversarial patches are limited in the following ways:

\begin{description}
	\item[Location] Only predictions around the patch itself are explicitly affected. This limits where objects can be made to `appear' in a scene. For example, a patch cannot make a plane appear in the sky and it is difficult to put a patch in the middle of a busy road. Furthermore, patches in noticeable areas can raise suspicion (e.g., a stop sign with a colorful patch on it).
	\item[Interpretation] Existing patches do not explicitly alter the shape or layout of a scene's perceived semantics. Changes to these semantics can be used to guide behaviors (e.g., drive a car off the road \cite{teichmann2018multinet} or change a head count \cite{he2020cpspnet}) and has wide implications on tasks such as surveillance \cite{iglovikov2017satellite,Artifici74:online} and medical screening \cite{prajna2021survey} among others.
\end{description}

In this paper we identify a new type of attack which we call a \textbf{R}emote \textbf{A}dversarial \textbf{P}atch (\textbf{RAP}). A RAP is an adversarial patch which can alter an image's perceived semantics from a remote location in the image. Our implementation of a RAP (IPatch) can be placed anywhere in the field of view and alter the predictions of nearly any predetermined location within the same view. This is demonstrated in Fig. \ref{fig:targeted} where an attacker has crafted an IPatch which causes a segmentation model to think that there is pavement (a sidewalk) in the middle of the road. Moreover, this adversarial attack is robust because the same patch works on different images using different positions and scales. Therefore, this attack more flexible and more covert than previous approaches. later in section \ref{sec:threat} we discuss the attack model further.

\begin{figure}[t]
	\centering
	\includegraphics[width=\columnwidth]{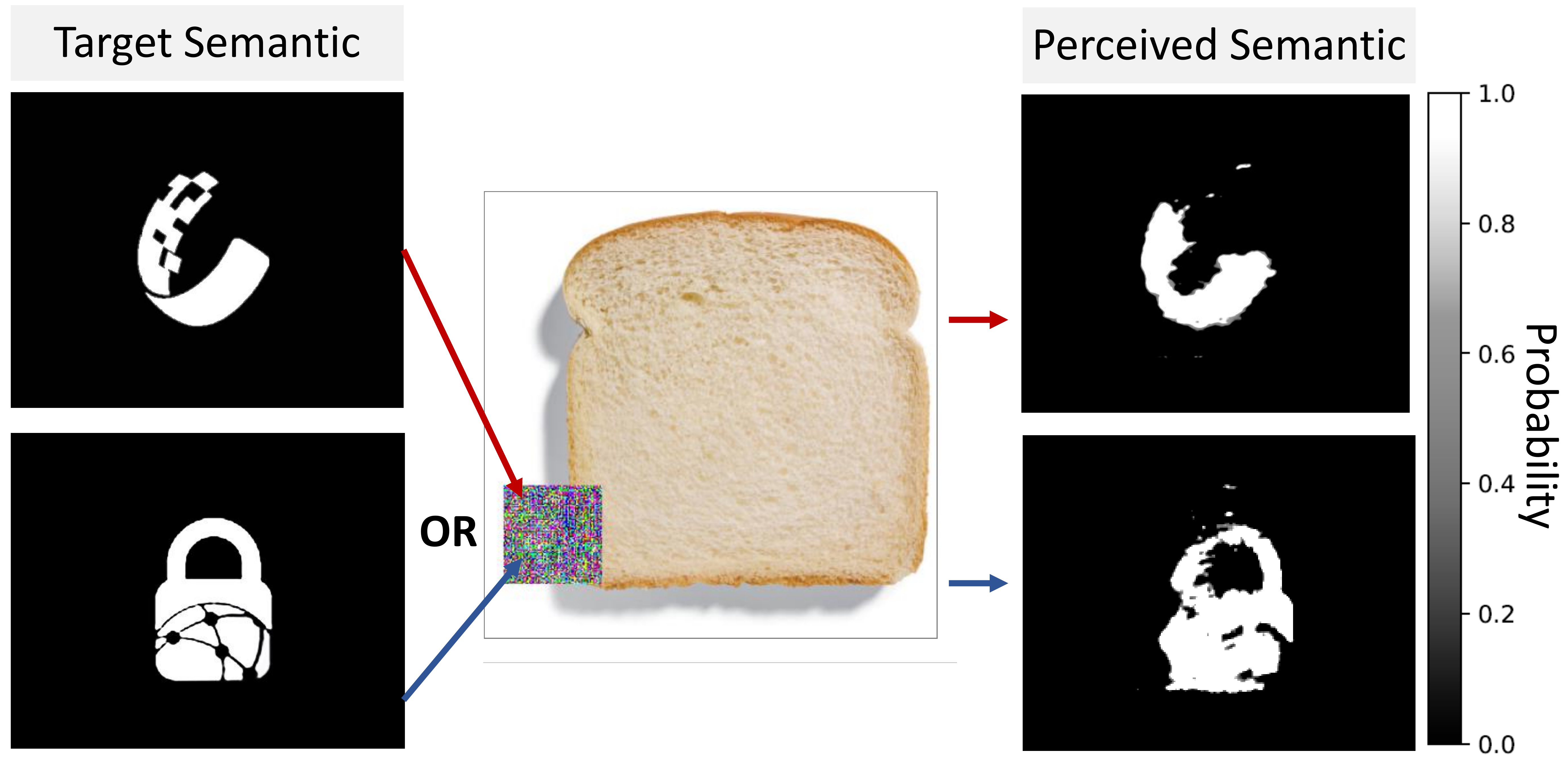}
	\caption{An example of how an IPatch can change the semantics of an image to an arbitrary shape. Here, a street view segmentation model is convinced that the slice of bread is a tree in the shape of the USENIX logo (top) and NDSS logo (bottom).}
	\label{fig:arbitrary}
\end{figure}

Since the IPatch can alter an image's perceived semantics, and attacker can craft patches which cause these models to see objects of arbitrary shapes and classes. For example, in Fig. \ref{fig:arbitrary} a street view segmentation model is convinced that a slice of bread is a tree shaped like the USENIX security symposium logo (top) or the NDSS logo (bottom). This is possible because semantic models rely on global and contextual features to parse an image. However, an object and its contextual information can be very far apart in $x$. For example, consider an image with a boat next to the water. Here, the water will boost the confidence of the boat's classification even though the boat is not in the water. The IPatch exploits these correlations by masquerading as these contextual features.

Creating a robust RAP is more challenging than existing adversarial patches. This is because the content of $x$ directly affects the leverage of the patch. For example, an IPatch cannot make a segmentation model perceive remote semantics on a blank image. However, to create a robust patch, we must be able to generalize to different images which have not been seen before. To overcome these challenges, we (1) use an incremental training strategy to slowly increase the entropy of the expectation over transformation (EoT) objective and (2) use Kullback-Leibler divergence loss to help the optimizer leverage and exploit the contextual relationships. 

In this paper, we focus on the use of IPatches as a RAP against semantic segmentation models. We also demonstrate that the same technique can be applied to object detectors, such as YOLO, as well. To evaluate the IPatch, we train 37 segmentation models using 8 different encoders and 5 state-of-the-art architectures. In our evaluations, we focus on the autonomous car scenario \cite{Autopilo49:online,siam2018comparative}, and perform rigorous tests to determine the limitations and capabilities of the attack. On the top 4 classes, we found that the attack works up to 93\% of the time on average, depending on the victim's model. We also found that all of the segmentation models are susceptible to the attack, where the most susceptible architectures were the \texttt{FPN} and \texttt{Unet++} and the least susceptible architecture was the \texttt{PSPNet}. Finally, even if the attacker does not have the same architecture as the victim, we found that without any additional training effort, an IPatch trained on one architecture works on others with an attack success rate of up to 25.3\%.


The contributions of this paper are as follows:
\begin{itemize}[itemsep=0em]
	\item We introduce a new class of adversarial patches (RAP) which can manipulate a scene's interpretation remotely and explicitly. This type of attack not only has significant implications on the security of autonomous vehicles, but also on a wide range of semantic-based applications such as medical scan analysis, surveillance, and robotics (section \ref{sec:threat}).
	\item We present a training framework which enables the creation of a robust RAP (IPatch) by incrementally increasing the training entropy. Without this strategy, the entropy starts too high which makes it difficult to converge on some learning objectives, especially given large patch transformations on scale, shift, and so on (section \ref{sec:ipatch}). 
	\item We provide an in-depth evaluation of the patch used as a remote adversarial attack against road segmentation models (section \ref{sec:eval}). We show that the attack is robust, universal (works on unseen images sampled from the same distribution), and has transferability (works across multiple models). We also provide initial results which demonstrate that the attack works on object detectors as well (specifically YOLOv3). 
	\item We identify the attack's limitations and provide insight as to why this attack can alter the perception of remote regions in an image. Building on these observations, we suggest countermeasures and directions for future work (section \ref{sec:dicussion}).
	\item To the best of our knowledge, this the first adversarial patch demonstrated on segmentation models (section \ref{sec:relatedworks}).
\end{itemize}

To reproduce our results, the reader can access the code and models used in this paper online. \footnote{The code and models will be made available here \url{https://github.com/ymirsky/IPatch}}

\begin{figure*}[t]
	\centering
	\includegraphics[width=\textwidth]{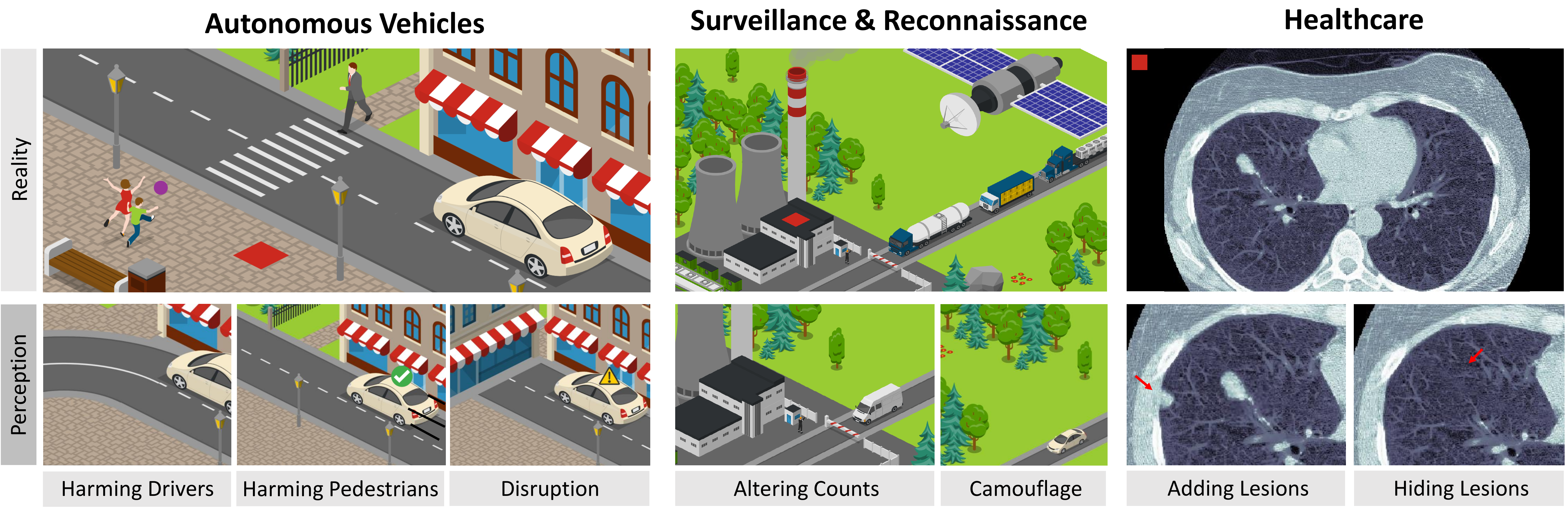}
	\caption{Some illustrative examples showing how a remote adversarial patch can be used by an attacker.}
	\label{fig:attackmodel}
\end{figure*}

\section{Related Works}\label{sec:relatedworks}
Soon after the popularization of deep learning, researchers demonstrated that DNNs can be exploited using adversarial examples \cite{biggio2018wild}. In 2014 it was shown how an attacker can alter an image-classification model's predictions by adding an imperceivable amount of noise to the input image \cite{szegedy2013intriguing,goodfellow2014explaining,nguyen2015deep}. Initially, these attacks were impractical to perform in a real environment since every combination of lighting, camera noise, and perspective would require a different adversarial perturbation \cite{luo2015foveation,lu2017no}. However, in 2017 the authors of \cite{athalye2018synthesizing} showed that an adversary can consider these distortions while generating the adversarial example in a process called Expectation over Transformation (EoT). Using this method, the authors were able to generate robust adversarial samples which can be deployed in the real world. 
In the same year, the authors of \cite{brown2017adversarial} used EoT to create adversarial patches. Their adversarial patches were designed to fool image-classifiers (single-object detection models).

Later in 2018, the authors of \cite{song2018physical} developed an adversarial patch that works on object detection models (multi-object detection models).
More recently, researchers have proposed patches which can remove objects which wear the patch \cite{liu2018dpatch,zhao2019seeing,	thys2019fooling,wu2020making,huang2020universal,den2020adversarial,li2020adaptive} and patches which can perform denial of service (DoS) attacks by corrupting a scene's interpretation \cite{liu2018dpatch,lee2019physical}. 

In Table \ref{tab:relworks}, we summarize the related works on adversarial examples against image segmentation and object detection models (the domain of the proposed attack). In general, the attack goals of these papers are either add/change an object in the scene or to remove all objects altogether (DoS). The methods which add adversarial perturbations (noise) can change the semantics of an image at any location \cite{hendrik2017universal,fischer2017adversarial,arnab2018robustness,ozbulak2019impact,kang2020adversarial}, but they cannot be deployed in the real world since they are applied directly to an image itself.  Currently, there no patches for image segmentation models, and the patches for object detection models only affect the prediction around the patch itself. The exception are patches which perform DoS attacks by removing/corrupting all objects detected in the scene like \cite{liu2018dpatch,lee2019physical}.

Therefore, to the best of our knowledge, the attack which we introduce is the first RAP, and (1) the only method which can add, change, or remove objects in a scene \textit{remotely} (far from the location of the patch itself), (2) the first adversarial patch proposed for segmentation networks, and (3) the first adversarial patch which can cause a model to perceive custom semantic shapes.

\bgroup

\begin{table}[t]
	\resizebox{\columnwidth}{!}{%
		\begin{tabular}{@{}ccc|cc|cc|cc|cc|@{}}
			&  &  & \multicolumn{2}{c|}{\textbf{Type}} & \multicolumn{2}{c|}{\textbf{Attack Goal}} & \multicolumn{2}{c|}{\textbf{Deployment}} & \multicolumn{2}{c|}{\textbf{Flexibility}} \\
			&  &  & \rotatebox[origin=c]{90}{Patch \hspace{6.5em}} & \rotatebox[origin=c]{90}{Noise\hspace{6.5em}} & \rotatebox[origin=c]{90}{Add/Change\hspace{4em}} & \rotatebox[origin=c]{90}{Hide\hspace{6.5em}} & \rotatebox[origin=c]{90}{Applied to image\hspace{2em}} & \cellcolor[HTML]{EFEFEF}\rotatebox[origin=c]{90}{Physically deployed\hspace{1em}} & \rotatebox[origin=c]{90}{Must be near/on target} & \cellcolor[HTML]{EFEFEF}\rotatebox[origin=c]{90}{Can be placed anywhere} \\ \midrule \midrule
			& 2018 & \cite{song2018physical} & $\bullet$ &  & $\bullet$ &  &  & \cellcolor[HTML]{EFEFEF}$\bullet$ & $\bullet$ & \cellcolor[HTML]{EFEFEF} \\
			& 2018 & \cite{chen2018shapeshifter} & $\bullet$ &  & $\bullet$ &  &  & \cellcolor[HTML]{EFEFEF}$\bullet$ & $\bullet$ & \cellcolor[HTML]{EFEFEF} \\
			& 2018 & \cite{liu2018dpatch} & $\bullet$ &  & $\bullet$ & $\bullet$ &  & \cellcolor[HTML]{EFEFEF}$\bullet$ & $\bullet$ & \cellcolor[HTML]{EFEFEF}$\bullet$* \\
			& 2018 & \cite{sitawarin2018darts} & $\bullet$ &  &  & $\bullet$ &  & \cellcolor[HTML]{EFEFEF}$\bullet$ & $\bullet$ & \cellcolor[HTML]{EFEFEF} \\
			& 2018 & \cite{wei2018transferable} &  & $\bullet$ & $\bullet$ &  & $\bullet$ & \cellcolor[HTML]{EFEFEF} & $\bullet$ & \cellcolor[HTML]{EFEFEF} \\
			& 2019 & \cite{zhao2019seeing} & $\bullet$ &  &  & $\bullet$ &  & \cellcolor[HTML]{EFEFEF}$\bullet$ & $\bullet$ & \cellcolor[HTML]{EFEFEF} \\
			& 2019 & \cite{thys2019fooling} & $\bullet$ &  &  & $\bullet$ &  & \cellcolor[HTML]{EFEFEF}$\bullet$ & $\bullet$ & \cellcolor[HTML]{EFEFEF} \\
			& 2019 & \cite{lee2019physical} & $\bullet$ &  &  & $\bullet$ &  & \cellcolor[HTML]{EFEFEF}$\bullet$ &  & \cellcolor[HTML]{EFEFEF}$\bullet$* \\
			& 2020 & \cite{wu2020making} & $\bullet$ &  &  & $\bullet$ &  & \cellcolor[HTML]{EFEFEF}$\bullet$ & $\bullet$ & \cellcolor[HTML]{EFEFEF} \\
			& 2020 & \cite{hoory2020dynamic} & $\bullet$ &  & $\bullet$ &  &  & \cellcolor[HTML]{EFEFEF}$\bullet$ & $\bullet$ & \cellcolor[HTML]{EFEFEF} \\
			& 2020 & \cite{zhao2020object} &  & $\bullet$ &  & $\bullet$ & $\bullet$ & \cellcolor[HTML]{EFEFEF} & $\bullet$ & \cellcolor[HTML]{EFEFEF} \\
			& 2020 & \cite{huang2020universal} & $\bullet$ &  &  & $\bullet$ &  & \cellcolor[HTML]{EFEFEF}$\bullet$ & $\bullet$ & \cellcolor[HTML]{EFEFEF} \\
			& 2020 & \cite{chow2020adversarial} &  & $\bullet$ & $\bullet$ & $\bullet$ & $\bullet$ & \cellcolor[HTML]{EFEFEF} & $\bullet$ & \cellcolor[HTML]{EFEFEF} \\
			& 2020 & \cite{den2020adversarial} & $\bullet$ &  &  & $\bullet$ &  & \cellcolor[HTML]{EFEFEF}$\bullet$ & $\bullet$ & \cellcolor[HTML]{EFEFEF} \\
			& 2020 & \cite{zhang2020contextual} &  & $\bullet$ & $\bullet$ &  & $\bullet$ & \cellcolor[HTML]{EFEFEF} & $\bullet$ & \cellcolor[HTML]{EFEFEF} \\
			& 2020 & \cite{zolfi2020translucent} &  & $\bullet$ &  & $\bullet$ &  & \cellcolor[HTML]{EFEFEF}$\bullet$ & $\bullet$ & \cellcolor[HTML]{EFEFEF} \\
			& 2020 & \cite{li2020fa} &  & $\bullet$ & $\bullet$ &  & $\bullet$ & \cellcolor[HTML]{EFEFEF} & $\bullet$ & \cellcolor[HTML]{EFEFEF} \\
			& 2020 & \cite{li2020adaptive} & $\bullet$ &  &  & $\bullet$ &  & \cellcolor[HTML]{EFEFEF}$\bullet$ & $\bullet$ & \cellcolor[HTML]{EFEFEF} \\
			\multirow{-20}{*}{\rotatebox[origin=c]{90}{Object Detection}} & 2020 & \cite{den2020adversarial} &  & $\bullet$ & $\bullet$ &  & $\bullet$ & \cellcolor[HTML]{EFEFEF} & $\bullet$ & \cellcolor[HTML]{EFEFEF} \\ \midrule
			& 2017 & \cite{fischer2017adversarial} &  & $\bullet$ &  & $\bullet$ & $\bullet$ & \cellcolor[HTML]{EFEFEF} & $\bullet$ & \cellcolor[HTML]{EFEFEF} \\
			& 2017 & \cite{hendrik2017universal} &  & $\bullet$ &  & $\bullet$ & $\bullet$ & \cellcolor[HTML]{EFEFEF} & $\bullet$ & \cellcolor[HTML]{EFEFEF} \\
			& 2018 & \cite{arnab2018robustness} &  & $\bullet$ & $\bullet$ &  & $\bullet$ & \cellcolor[HTML]{EFEFEF} & $\bullet$ & \cellcolor[HTML]{EFEFEF} \\
			& 2020 & \cite{kang2020adversarial} &  & $\bullet$ & $\bullet$ &  & $\bullet$ & \cellcolor[HTML]{EFEFEF} & $\bullet$ & \cellcolor[HTML]{EFEFEF} \\
			\multirow{-5}{*}{\rotatebox[origin=c]{90}{Img. Segment.}} & 2019 & \cite{ozbulak2019impact} &  & $\bullet$ & $\bullet$ &  & $\bullet$ & \cellcolor[HTML]{EFEFEF} & $\bullet$ & \cellcolor[HTML]{EFEFEF} \\ \midrule
			& 2017 & \cite{xie2017adversarial} &  & $\bullet$ & $\bullet$ & $\bullet$ & $\bullet$ & \cellcolor[HTML]{EFEFEF} & $\bullet$ & \cellcolor[HTML]{EFEFEF} \\
			\multirow{-2}{*}{\rotatebox[origin=c]{90}{Both}} & \cellcolor[HTML]{C0C0C0} & \cellcolor[HTML]{C0C0C0}\textbf{IPatch} & \cellcolor[HTML]{C0C0C0}$\bullet$ & \cellcolor[HTML]{C0C0C0} & \cellcolor[HTML]{C0C0C0}$\bullet$ & \cellcolor[HTML]{C0C0C0}$\bullet$ & \cellcolor[HTML]{C0C0C0} & \cellcolor[HTML]{C0C0C0}$\bullet$ & \cellcolor[HTML]{C0C0C0} & \cellcolor[HTML]{C0C0C0}$\bullet$ \\ \midrule\bottomrule
			\multicolumn{11}{l}{*Patch can be anywhere only when used to hide all objects in the scene (DoS)}
		\end{tabular}%
	}
	\caption{Related works on adversarial examples which target image segmentation and object detection.}
	\label{tab:relworks}
\end{table}
\egroup

\section{Threat Model}\label{sec:threat}

\noindent\textbf{The Vulnerability.} The vulnerability which this paper introduces is that semantic models, such as image segmentation models, utilize global and contextual features in an image to improve their predictive capabilities. However, these dependencies expose channels which can an attacker can exploit to change the interpretation of an image from one remote location to another. 

\noindent\textbf{The Attack Scenario.}
In this work we will focus on the remote adversarial attack scenario. In this attack scenario, Alice has an application which uses the image segmentation model $M$. Mallory wants $M$ to predict a specific class at a specific location $L$, while looking at a certain scene. To accomplish this, Mallory needs a training set of 1 or more images and a segmentation model to work with. 

For the training set, Mallory has two options: (1) obtain images similar to those used to train $M$, or (2) take pictures of the target scene. For the model, Mallory can either follow a white-box or black-box approach: In a white box-approach, Mallory obtains a copy of $M$ to achieve the most accurate results. The white-box approach is a common assumption for adversarial patches. Alternatively, Mallory can follow a black-box approach and train a surrogate model $M'$ on a similar dataset used to train $M$. Although the black box approach performs worse, we have found that there is some transferability between a patch trained on one model and then used against another (section \ref{sec:eval}). Finally, Mallory generates an IPatch $P$ which targets $L$ using $X$ and $M$.

\noindent\textbf{Motivation.} 
There are several reasons why an adversary would want to use an IPatch over an ordinary adversarial patch (illustrated in Fig. \ref{fig:attackmodel}:
\begin{description}[itemsep=0em]
	\item[Stealth] The attacker may want to place the patch in a less obvious place so it won't be removed or noticed by the victim. For example, a sticker on a stop sign is anomalous and can be contextually identified as malicious \cite{sitawarin2018darts} but a sticker on a nearby billboard is less obvious. Another example, is in the domain of medicine where segmentation models are used to highlight and identify different lesions such as tumors. Here, an attacker can't put a patch in the image in the location of the lesion since it would be an obvious attack. However, the attack could be trigger remotely by placing a dark RAP in the dark space of a scan where it is common to have noise, or in a location of the scan which is not under investigation (e.g., the first few slices on the z-axis). For motivations why an attacker would want to target medical scans, see \cite{236284}.
	
	\item[Practicality] The attacker may want to generate an object or semantic illusion in a location which is hard to reach or impractical to place a patch on it. For example, in the sky region, on the back of an arbitrary car on the freeway. 	
	
	\item[Flexibility] The attacker may need to craft or alter specific semantics for a scene. For example, many works show how image segmentation can be used to identify homes, roads and resources from satellite and drone footage \cite{audebert2017segment}. Here an attacker can feed false intel by hiding or increasing the number of structures, people, and resources before it can be investigated manually. 
\end{description}

Overall, the IPatch attack is more flexible and enables more attack vectors than location-based patches (e.g., \cite{song2018physical,den2020adversarial}). However, it is significantly more challenging to generate an IPatch. Therefore, its flexibility comes with a trade-off in terms of attack performance.

\section{Making an IPatch}\label{sec:ipatch}

In this section we first provide an overview of how image segmentation models work. Then we present our approach on how to create an IPatch.

\subsection{Technical Background}\label{subsec:technical}
\begin{figure}[t]
	\centering
	\includegraphics[width=.9\columnwidth]{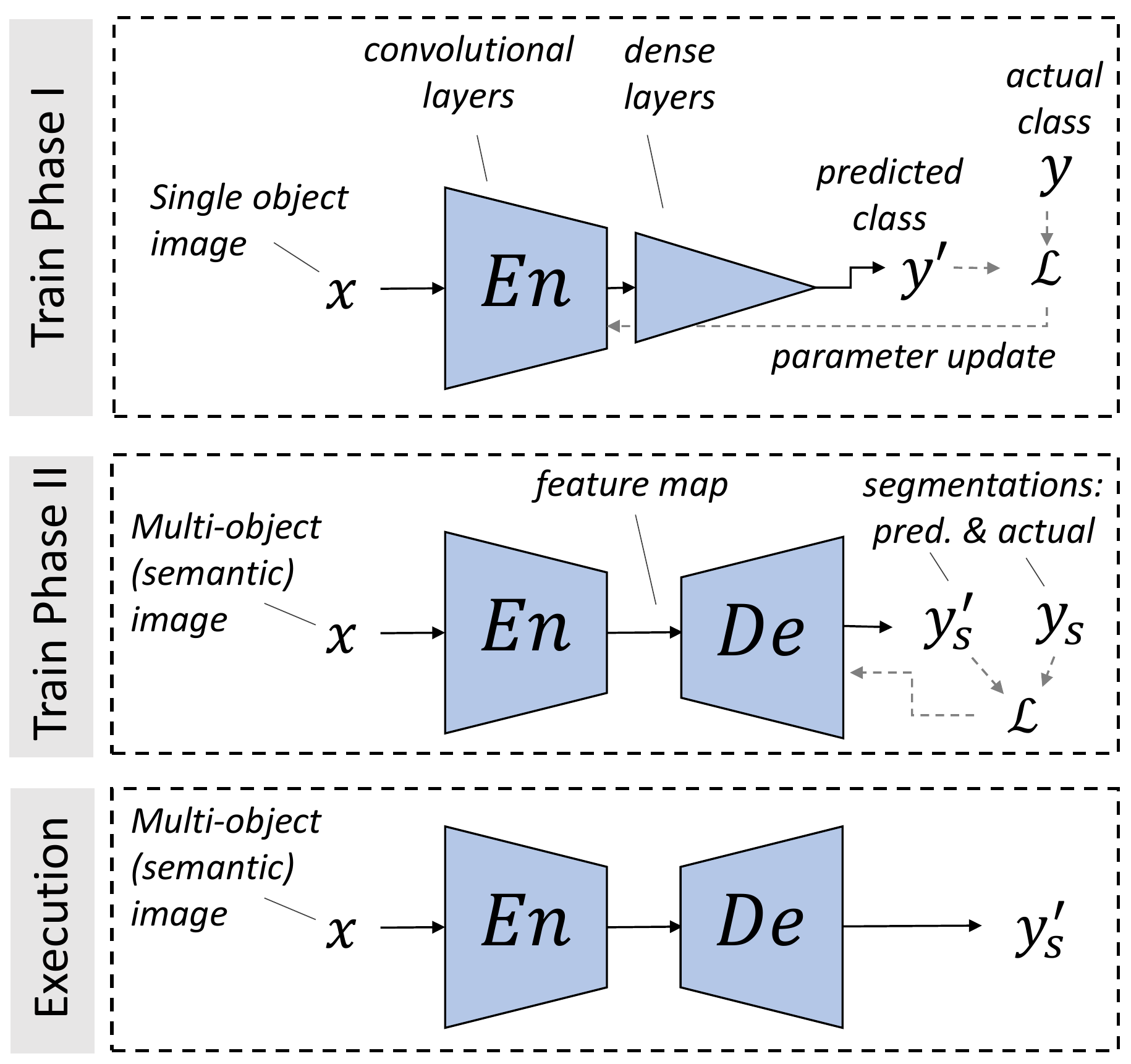}
	\caption{A basic schematic showing how a common Image Segmentation model is trained and executed.}
	\label{fig:segmodel}
\end{figure}

There are a wide variety of deep learning models for image segmentation \cite{garcia2018survey,minaee2020image}. The most common form involves an encoder $En$ and decoder $De$ such that
\begin{equation}
S(x) = De(En(x))
\end{equation}
illustrated in Fig. \ref{fig:segmodel}. The objective of a segmentation model is to take an $N$-by-$M$ image ($s$) with 1-3 color channels and predict an $N$-by-$M$-by-$C$ probability mapping ($y_s$). The output $y_s$ can be mapped directly to the pixels of $x$ such that $y_s[i,j,k]$ is the probability that pixel $x[i,j]$ belongs to the $k$-th class (among the $C$ possible classes). 

To train $S$, the common approach is to follow two phases: In the first phase, the encoder network is trained as an image-classifier on a large image dataset in a supervised manner (i.e., where each image $x$ is associated with a label $y$). Note that the classifier's task is to predict a single class for the entire image (e.g., $x$ is a dog). After training the classifier, we discard the dense layers at the end of the network (used to predict $y$) and retain the convolutional layers at the front of $En$. In this way, we can use the feature mapping learned by the classifier to perform image segmentation.  In the second phase, the decoder architecture $De$ is added on and $S$ is trained end-to-end. Often, the weights of $En$ are locked during this phase, and we do the same in this paper.

One reason why the encoder-decoder approach is so popular, is because obtaining a labeled segmentation ground truth $y_s$ is significantly more challenging than for image classification $y$ (massive datasets for classification exist and new datasets can be crowd sourced as well). Therefore, by using a pre-trained encoder, far fewer examples of segmentations are needed to achieve quality results.

To train $S$, a differentiable loss function $\mathcal{L}$ is used to compare the model's predicted output $y_s'$ to the ground truth $y_s$ in order to perform backpropagation and update network's weights. There are many loss functions used in for segmentation. One common approach is to simply apply the binary cross entropy loss ($\mathcal{L}_{CE}$) since $S$ is essentially trying to solve a multi-class classification problem. However, $\mathcal{L}_{CE}$ does not consider whether a pixel is on the boundary or not so results tends to be blurry and be biased to large segments such as backgrounds \cite{deng2018learning}. To counter this issue, in 2016 the authors of \cite{milletari2016v} proposed using Dice loss ($\mathcal{L}_{D}$) for medical image segmentation, and it has since been considered a state-of-the-art approach. The Dice loss is defined as
\begin{equation}
\mathcal{L}_{D}(x,y) = \frac{2 \sum_{i}^{N} x_i y_i}{\sum_{i}^{N} x^2_i + \sum_{i}^{N} y^2_i}
\end{equation}
We use $\mathcal{L}_{D}$ to train all of the image segmentation models in this paper.

When selecting the encoder's model, there are a wide variety of options. Some include \texttt{ResNext}, \texttt{DenseNet}, \texttt{xception}, \texttt{EfficientNet}, \texttt{MobileNet}, \texttt{DPN}, \texttt{VGG}, and variations thereof. However, regarding the decoder's architecture, there are several which are considered state-of-the-art. Many of them utilize a `feature pyramid' approach and skip connections to identify features at multiple scales, or an autoencoder (encoder decoder pair) to encode and extract the semantics. We will now briefly describe the five architectures used in this paper:

\begin{description}[itemsep=0em]
	\item[Unet++ \cite{zhou2018unet++}:] An autoencoder architecture which improves on its predecessor, the Unet. The encoder and decoder are connected through a series of nested dense skip connections which reduce the semantic gap between the feature maps of the two networks.
	
	\item[Linknet \cite{chaurasia2017linknet}:] An efficient autoencoder which passes spatial information across the network to avoid losing it in the encoder's compression.
	\item[FPN \cite{lin2017feature}:] A feature pyramid network which uses lateral connections across a fully convolution neural network (FCN) to utilize feature maps learned from multiple image scales.
	\item[PSPNet \cite{zhao2017pyramid}:] An FCN which uses a pyramid parsing module on different sub-region representations in order to better capture global category clues. The architecture won first place in multiple segmentation challenge contests. 
	\item[PAN \cite{li2018pyramid}:] A network which uses both pyramid and global attention mechanisms to capture spatial and global semantic information.
\end{description}

\subsection{Approach}\label{subsec:approach}

In a remote adversarial attack, the attacker wants a region around the location $L=(i,j)$ to be predicted as class $k$. To ensure that the optimizer does not waste energy on other semantics in the scene, we focus the effort to a region of operation. Let $m$ denote the region operation and let $t$ be the target pattern for that region. To capture $m$, we use an $N$-by-$M$-by-$C$ mask of zeros. To select $L$, a square or circle with a radius of $r$ pixels\footnote{For an $x$ with a dimension of  384x480, we found that a radius of 50 pixels empirically performs best when targeting region with a radius of 10 pixels.} around $L$ in $m$ is marked with ones along the $k$-th channel. To insert an object, we set $t=m$ since our objective is to change the probability of those pixels to one. To insert a custom shape (like in Fig. \ref{fig:arbitrary}) $t$ is set accordingly.

To generate a patch for the objective $(L,k)$, we follow the EoT approach similar to previous works \cite{brown2017adversarial,	song2018physical,liu2018dpatch}, but using our semantic masks. Concretely, we would like to find a patch $P$ which is trained to optimize the following objective function
\begin{equation}\label{eq:objective}
P=\arg\min_{\hat{P}}  \mathop{\mathbb{E}}_{x\sim X,l\sim \ell,s\sim S}[ S(A(x,\hat{P},l,s))\odot m - t]
\end{equation} 
where $X$ is a distribution of input images, $\ell$ is a distribution of patch locations, and $S$ is a set of scales to resize $\hat{P}$. The operator $A$ is the `Apply' procedure which takes the current $\hat{P}$ and inserts it into $x$ while sampling uniformly on the distributions $X$, $\ell$, and $S$. 

\noindent\textbf{Loss Function.} We experimented with many different loss functions on the CamVid dataset \cite{brostow2009semantic}: $\mathcal{L}_{CE}$,$\mathcal{L}_{D}$,$\mathcal{L}_1$,$\mathcal{L}_2$, and $\mathcal{L}_{KL}$ (Kullback–Leibler Divergence loss). Most of these loss functions took too long to converge or got stuck in local optima. Instead, we found that  $\mathcal{L}_1$ works best in emptier scenes (like Fig. \ref{fig:arbitrary}) and $\mathcal{L}_{KL}$ works best in busy scenes like those in CamVid. We believe the reason why $\mathcal{L}_{KL}$ performs well busy scenes is because it measures the relative entropy from one distribution to another. As a result, the optimizer had an easier time `leeching' nearby features and contexts in $x$ to match the goal in $t$.

\noindent\textbf{Creating a Robust RAP.}
In order to make an RAP which is robust to different transformations (scale and location), and universal to different images (not in the training set of the patch), we must use EoT. However, in some cases we found that the patch does not converge well when the range of ($X$, $\ell$, $S$) is large (i.e, large shifts, hundreds of images, etc). This is because (1) the IPatch leverages the variable contents of $x$ to impact $S(y_s')$ and (2) the placement of $P$ in $x$ affects the influence of $P$ on $S(y_s')$. 

For these cases, we propose an incremental training strategy where we gradually increase the placement radius of the patch $\ell$. Whenever the training has converged or a time limit has elapsed, we increase the radius by one pixel. We repeat this process until the entire dataset is covered. At the start of each epoch, we give the optimizer time to adjust by setting the learning rate to a fraction of its value and then slowly ramp it back up. A similar strategy can be applied to the other distributions, such as the number of images in $X$, the shift size, or the patch scale.
This strategy works well because we gradually increase the entropy, enabling the optimizer capture foundational concepts. It can also be viewed that at each epoch we are placing the gradient descent optimizer at a more advantageous position instead of a random starting point. 

\begin{figure}[t]
	\centering
	\includegraphics[width=\columnwidth]{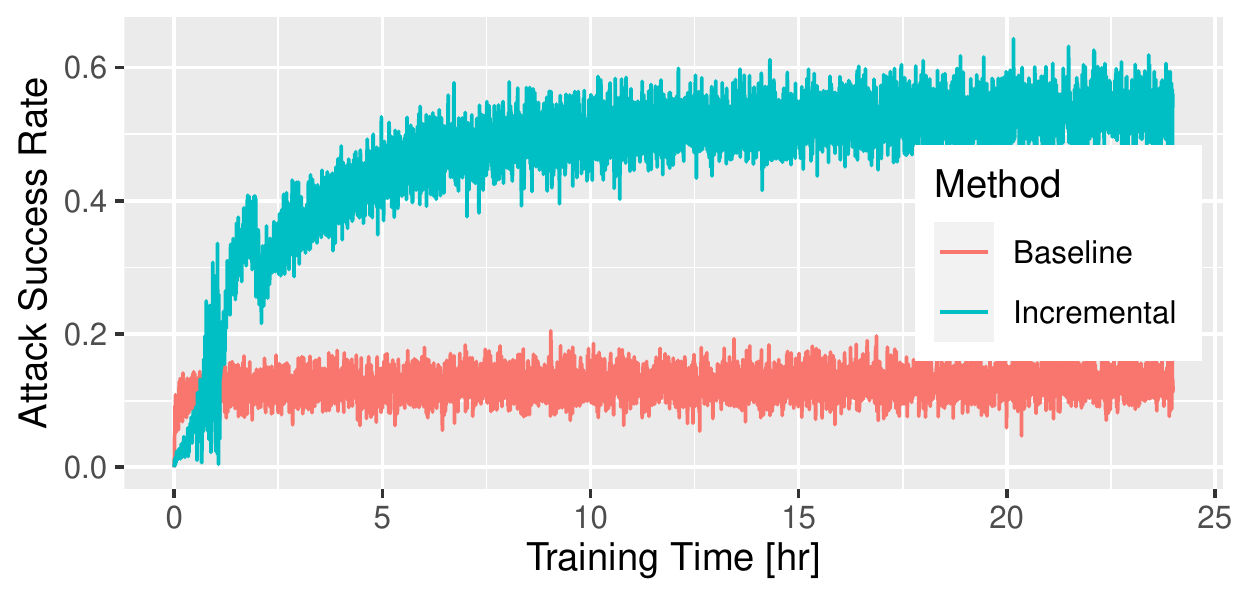}
	\caption{The benefit of using the proposed incremental training strategy (incrementing the patch placement radius) when training on the BDD100k dataset.}
	\label{fig:incResults}
\end{figure}

To demonstrate the value of the incremental strategy, we performed an experiment. We trained a RAP using 70k images from the BDD100k street segmentation dataset \cite{yu2018bdd100k}. The RAP was configured to make the center of th image perceived as the class `tree' when placed in \textit{any} location within the image (i.e., a placement radius of 500 pixels). The experiment was performed using (1) our incremental training strategy by increasing the placement radius up to the maximum radius and (2) the baseline approach of training the patch using the maximum radius from the start. For the incremental strategy, the placement radius was increased by one pixel whenever the attack success rate reached 25\%. 
In Fig. \ref{fig:incResults}, we plot the results from the experiment. We found that that our approach reaches the maximum radius after one hour and then exceeds the performance of the baseline shortly after by a margin of 35\%.

\begin{figure*}[t]
	\centering
	\includegraphics[width=\textwidth]{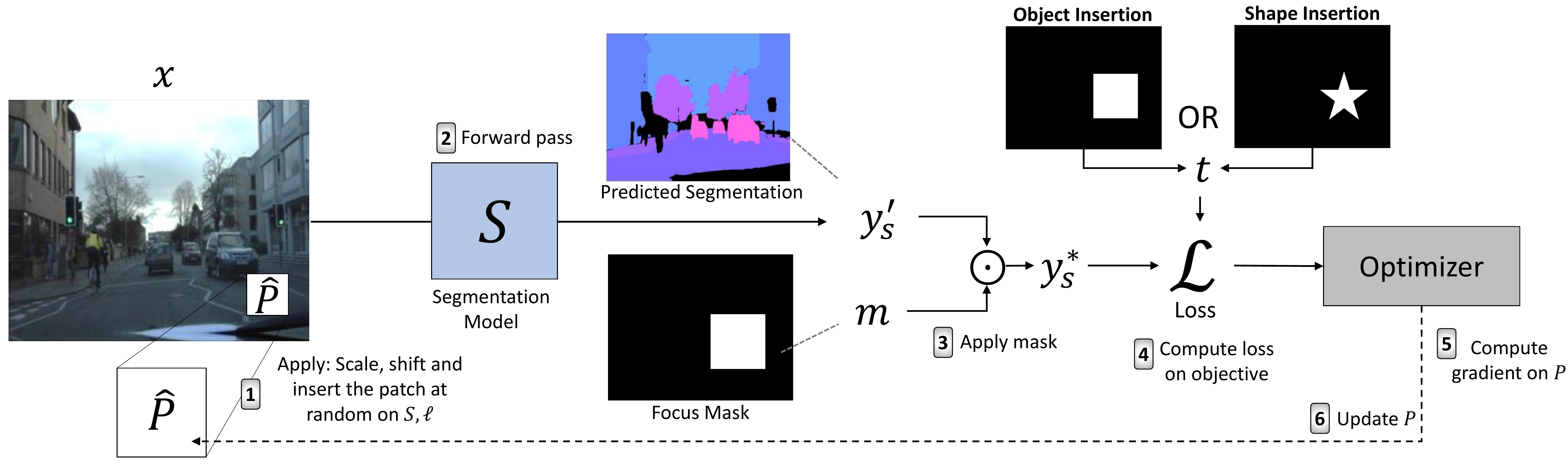}
	\caption{An overview of the IPatch training framework for creating a RAP. For simplicity, only one sample $x$ is shown, though in practice batches of images are used.}
	\label{fig:framework}
\end{figure*}

In summary, the training framework for creating an IPatch is as follows (illustrated in Fig. \ref{fig:framework}):
\begin{tcolorbox}[breakable,title=\textit{Training Procedure for an IPatch}]
	\small{
		Initialize $\hat{P}$ with random values and set its origin (default location in $x$) to be $o$.
		If \texttt{incremental}, then add one image to $X$. Otherwise, add all images to $X$.\\
		Repeat until $\hat{P}$ has converged on the entire dataset: 
		
		\begin{enumerate}[leftmargin=*]
			\item \textbf{Apply}: Draw a batch of samples from $X$. For each sample $x$ in the batch, perform a random transformation: scale down $\hat{P}$ and shift its location from origin $o$.
			
			\item\textbf{Forward pass}: Pass the batch through $S$ and obtain the segmentation maps (as a set of $y_s'$).
			
			\item\textbf{Apply mask}: Take the product of each $y_s'$ with the mask $m$ to omit irrelevant semantics.
			
			\item\textbf{Loss \& gradient}: Compute the loss $\mathcal{L}_{KL}(y_s^*,t)$ and use it to perform back propagation through $S$ to $\hat{P}$.
			
			\item[5-6.]\textbf{Update}:  Use gradient descent (e.g., Adamax) to update the values of $\hat{P}$.
			
			\item[7.] If \texttt{incremental} and the has time elapsed or training has converged, then increase the entropy (e.g., patch placement radius) and ramp the learning rate.
			
		\end{enumerate}
	}
\end{tcolorbox}

\section{Evaluation}\label{sec:eval}

To evaluate the IPatch as a RAP, we will focus our evaluation on the scenario of autonomous vehicles. The task of street view segmentation is challenging because the scenes are typically very busy with many layers, objects, and wide perspectives \cite{siam2018comparative}. Therefore attacking this application is will provide us with good insights into the IPatch's capabilities.

\noindent\textbf{Datasets.}
We use the CamVid dataset \cite{brostow2009semantic} to train our segmentation models and evaluate our adversarial patches. The CamVid dataset is a well-known benchmark dataset used for image segmentation. It contains 46,869 street view images with a resolution of 360x480 from the point of view of a car. The images are supplied with pixel-wise annotations which indicate the class of the corresponding content (e.g., car, building, etc). 
The dataset comes split into three partitions: train $D_{train}$, test $D_{test}$, and validation $D_{val}$. We use $D_{train}$ to train the segmentation models and the rest to train the patches. This way there is will be no bias on the images which we attack. The $D_{test}$ dataset is used to evaluate the influence of the patch's parameters (size and location) and to train robust patches with EoT. Finally, the $D_{val}$ dataset is use to validate that the robust patches work on unseen imagery.

\noindent\textbf{Segmentation Models.} 
In our evaluations we trained and attacked 37 different models which were combinations of 8 different encoders and 5 state-of-the-art segmentation architectures.\footnote{Every combination of encoder and architecture except for the architecture \texttt{PAN} which was incompatible with three of the encoders.} The encoders were the \texttt{vgg19}, \texttt{densenet121}, \texttt{efficientnet-b4}, \texttt{efficientnet-b7}, \texttt{mobilenet\_v2}, \texttt{resnext50\_32x4d}, \texttt{dpn68}, and \texttt{xception}. All of the models were obtained from the Torch library and were pretrained on the ImageNet Dataset \cite{deng2009imagenet}. For the architectures, we used the implementations\footnote{\url{https://github.com/qubvel/segmentation_models.pytorch}} of the state-of-the-art segmentation networks described in section \ref{subsec:technical}. The models were trained on $D_{train}$ for 100 epochs each, with a batch size of 8, learning rate of 1e-4, using Dice Loss $\mathcal{L}_{D}$ and an Adam optimizer.  Finally, to increase the training set size and improve generalization, we performed data augmentation. The augmentations were: flip, shift, crop, blur, sharpen and change perspective, brightness, and gamma. 

\noindent\textbf{The Experiments.}
We performed three experiments:
\begin{description}
	\item[EXP1:] In this experiment we investigate the influence which a patch's size and location have on the attack performance. We also investigate the influence of the remote target's size and location. Here patches are crafted to target individual images. Therefore, the results of this experiment also tell us how well the attack performs on static images. 
	\item[EXP2:] To use this attack in the wild, the patch must work under various transformations and in new scenes. This experiment evaluates the attack's robustness by (1) training the patches with EoT according to (\ref{eq:objective}), and by (2) measuring the performance of these patches on new images (unseen during training). 
	\item[EXP3:] To get an idea of the vulnerability's prevalence, we attack 32 different segmentation models and measure their performance. To evaluate the case where the attacker has no information on the model, we take the robust patches trained in EXP2 and use them on the other 36 models to measure the attack's transferability. 
\end{description} 

When measuring attack performance, we omit all cases where the targeted region already contains the target class. For all of the experiments, we trained on an NVIDIA Titan RTX with 24GB of RAM. For the optimizer, we experimented on a variety of options in the Torch library. We found that the Adamax optimizer works best on the CamVid Dataset.

\subsection{EXP1: The Impact of Size and Location}\label{subsec:exp1}
The purpose of this experiment is to see how the size and locations of a patch and its target affect the attack's performance. In this experiment, we craft patches which target a single image. Later in section \ref{subsec:exp2} we evaluate multi-image `robust' patches. 

\noindent\textbf{Experiment Setup.} For EXP1 we attacked the \texttt{efficientnet-b7\_FPN} model since it performed best on the CamVid dataset. A list the evaluations and parameters used in EXP1 can be found in Table \ref{tab:exp1}. For each of these parameters, we varied their values while locking the rest to measure their influence. This was repeated for each of the model's top six performing classes. Due to time restrictions,\footnote{Each of these experiments on takes 3-5 days on a NVIDIA Titan RTX with 24GB RAM.} we only used the entirety of $D_{test}$ for the fixed parameter experiment. For the other experiments, we used 20 random images from $D_{test}$. 

The training procedure was as follows: For each patch, we used a learning rate of 2.5 and stopped the training after three minutes to ensure that each of the five experiments would take no more than 5 days. We note that in many cases, the patches were still converging so the results can be improved. Finally, we count a successful attack as any image with at least 80\% of $t$ marked by the model as the target class.

\begin{table}[t]
	\resizebox{\columnwidth}{!}{%
		\begin{tabular}{ccccc}
			\multicolumn{1}{c|}{} & \multicolumn{4}{c|}{\textbf{Parameters}} \\
			\multicolumn{1}{c|}{} & \multicolumn{2}{c|}{Patch $P$} & \multicolumn{2}{c|}{Target $t$} \\
			\multicolumn{1}{c|}{Experiment} & Size* $S$ & \multicolumn{1}{c|}{Location $\ell$} & Size $r$ & \multicolumn{1}{c|}{Location $L$} \\ \hline\hline
			\multicolumn{1}{c|}{\textbf{Locked}} & 80 & \multicolumn{1}{c|}{(370,270)} & 20 & \multicolumn{1}{c|}{(250,250)} \\
			\multicolumn{1}{c|}{\textbf{Patch Size}} & \textbf{30:130} & \multicolumn{1}{c|}{(370,270)} & 20 & \multicolumn{1}{c|}{(250,250)} \\
			\multicolumn{1}{c|}{\textbf{Patch Location}} & 80 & \multicolumn{1}{c|}{\textbf{**}} & 20 & \multicolumn{1}{c|}{(250,250)} \\
			\multicolumn{1}{c|}{\textbf{Target Size}} & 80 & \multicolumn{1}{c|}{(370,270)} & \textbf{20:280} & \multicolumn{1}{c|}{(250,250)} \\
			\multicolumn{1}{c|}{\textbf{Target Location}} & 80 & \multicolumn{1}{c|}{(370,270)} & 20 & \multicolumn{1}{c|}{***} \\ \hline\hline
		\end{tabular}%
	}
	\scriptsize *$P$ is scaled down to $S$ from 100x100. **On the diagonal from the image center to bottom right. ***8x6 grid over entire image
	\caption{The parameters used in EXP1. The values listed for `size' are both the height and width.}
	\label{tab:exp1}
\end{table}

\subsubsection{Performance with all Parameters Locked}

\begin{figure}[t]
	\centering
	\includegraphics[width=\columnwidth]{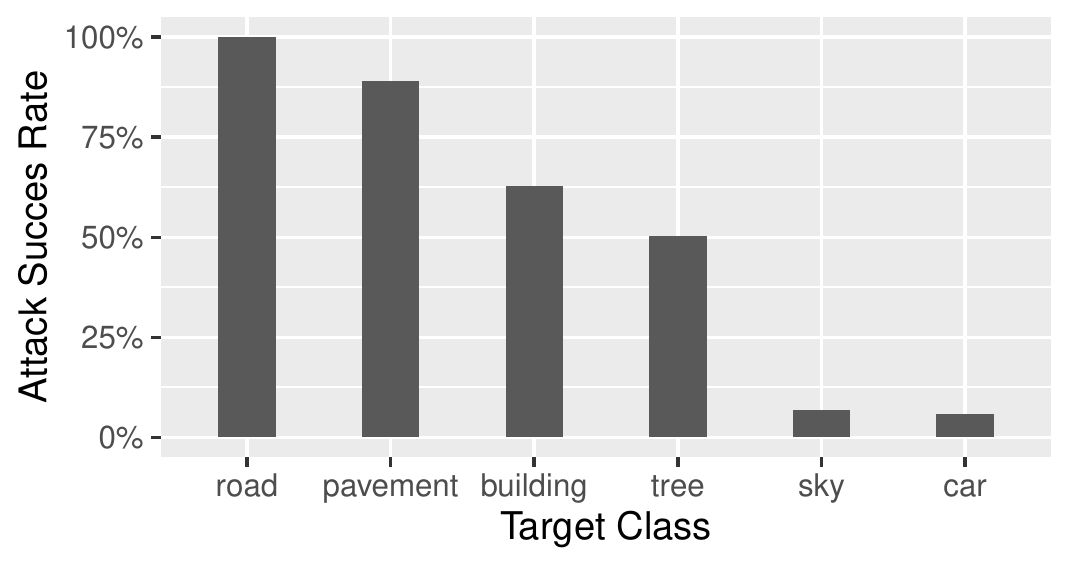}
	\includegraphics[width=\columnwidth]{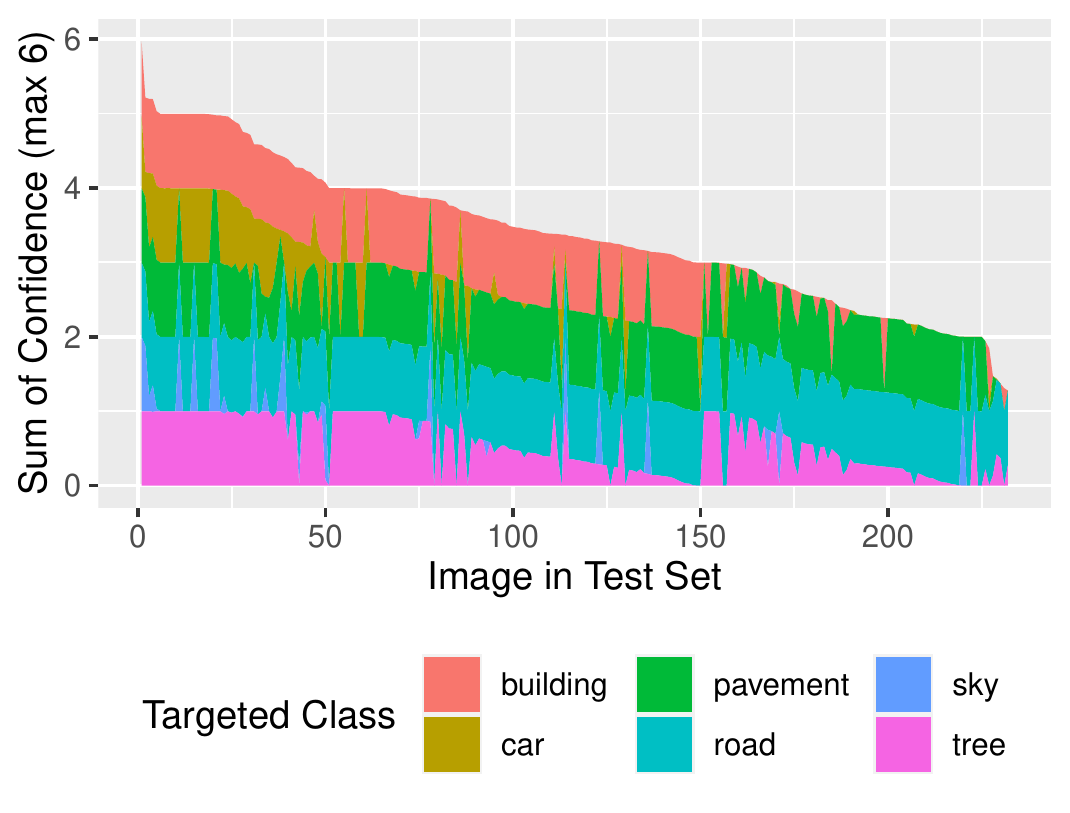}
	\caption{The attack performance with locked parameters. Top- The average attack performance per class. Bottom- the aggregated confidence of the model on each image, where the max confidence per class is one.}
	\label{fig:exp1a_1b}
\end{figure}

The results for the experiment, where the patch parameters are locked, can be found in Fig. \ref{fig:exp1a_1b}. The top of the figure shows that the attack has a greater impact on structural classes than others. This might be because these semantics have the largest regions CamVid dataset (i.e., are common). As a result, the patch is able to leverage these contexts better from one side of an image to another. For example, if there are is a row of buildings on one side of the road, then there is a higher probability that the other side will have one too. This kind of correlation is exploited by the patch. The road class out performs all the rest because the target $t$ in this experiment is in the center of the image, where the road is most commonly found (77\% of the images). However, the patch is able to successfully attack the classes of pavement, building, and tree at the same location, even though on the clean images, the model predicts  $1.4$\%, $0.3$\%, and $0$\% of them to have these classes respectively.

At the bottom of  Fig. \ref{fig:exp1a_1b} we can see the aggregated confidence of the model for each of the images in $D_{test}$. The plot shows that all of the images are susceptible to the attack for at least one of the target classes.

\subsubsection{Impact of the Patch Size}
\begin{figure}[t]
	\centering
	\includegraphics[width=\columnwidth]{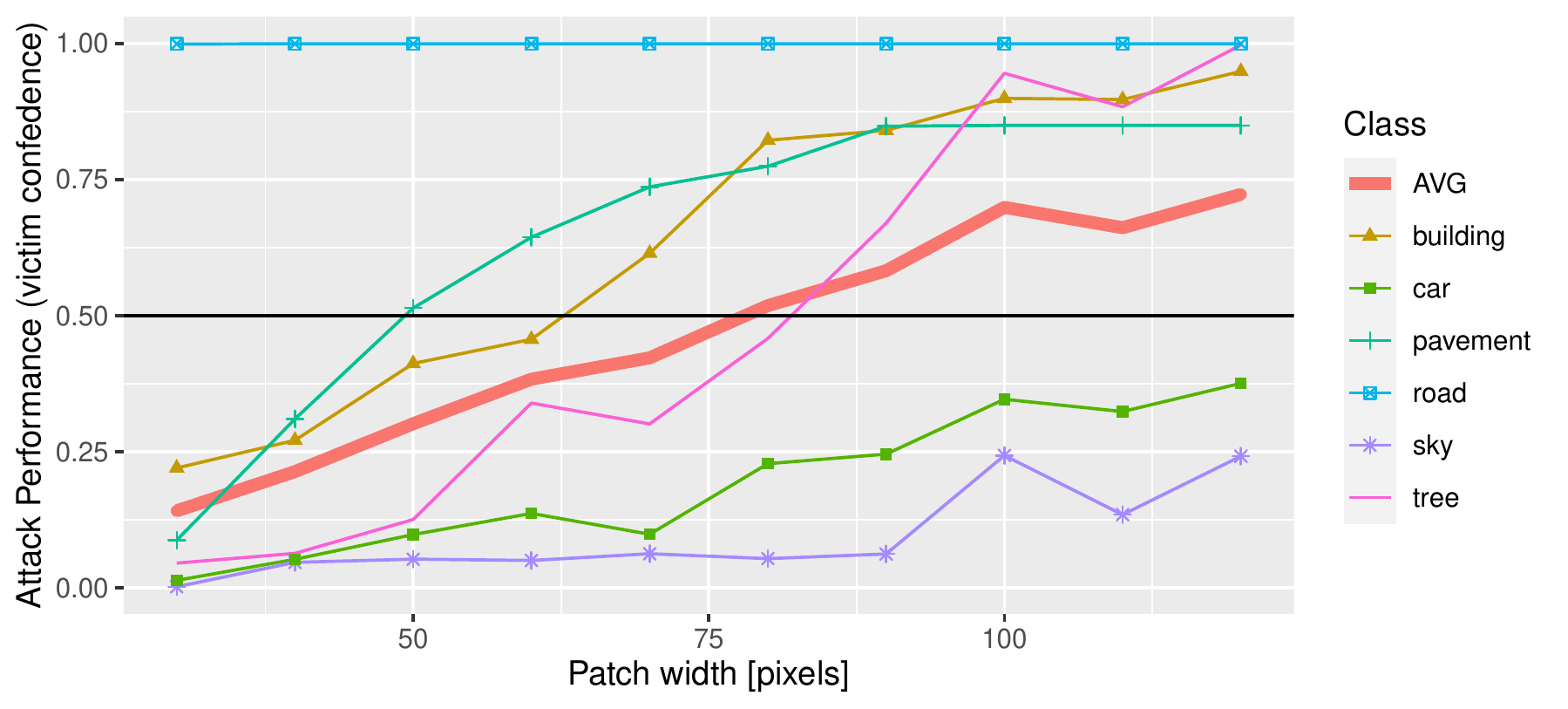}
	\caption{The average attack performance with different patch sizes.}
	\label{fig:exp2a_2c}
\end{figure}

Figure \ref{fig:exp2a_2c} plots the model's confidence over increasingly larger patch sizes. In the figure, we have marked 0.5 as the decision threshold which is the default for segmentation models. This is because segmentation models perform binary-classification on each pixel. As a result, the confidence scores per class are either close to zero or one, but not so much in between (as seen in Fig. \ref{fig:exp1a_1b}).

As expected, larger patches increase the attack success rate.  However, the trade off appears to be linear (captured by the average in red). What is meaningful about these results is that some classes excel with smaller patch sizes (e.g., pavement and building) while others require larger ones to succeed (e.g., tree). This is probably because some of the remote contextual semantics which the model considers cannot be compressed into small spaces when others can. Overall, we observe that the minimum patch size required to fool the model on a static image this size is about 60-75 pixels in width, and with a patch width of 100 pixels, nearly all attacks succeed.

\begin{figure*}[t]
	\centering
	\includegraphics[width=\textwidth]{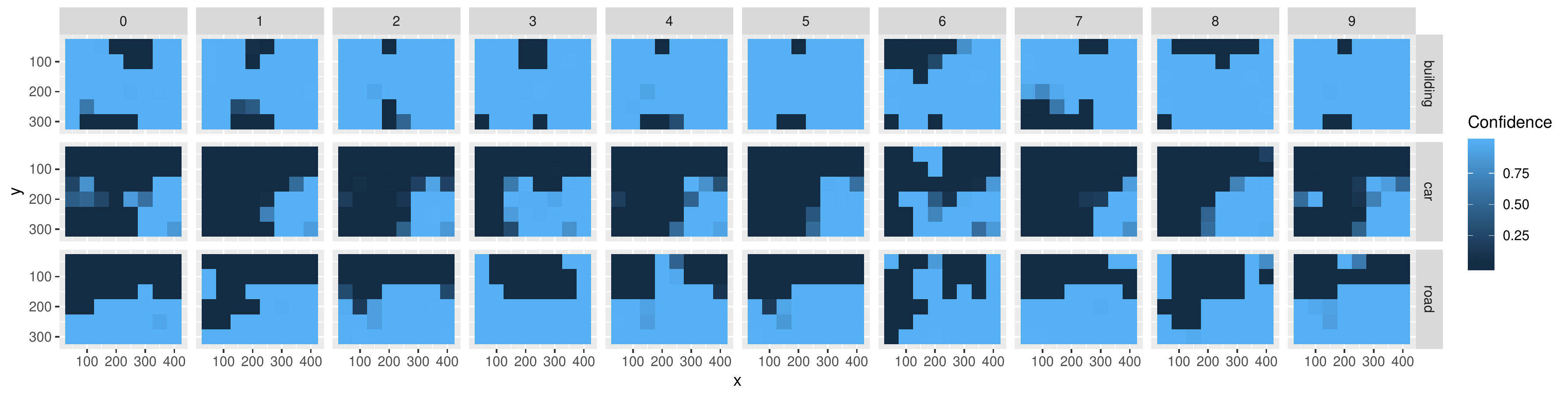}
	\caption{The remote attack performance on ten images. A region in the heat map indicates the performance of the attack when targeting that location.}
	\label{fig:exp5b}
\end{figure*}

\subsubsection{Impact of the Patch Location}\label{subsec:patch_location}
\begin{figure}[t]
	\centering
	\includegraphics[width=\columnwidth]{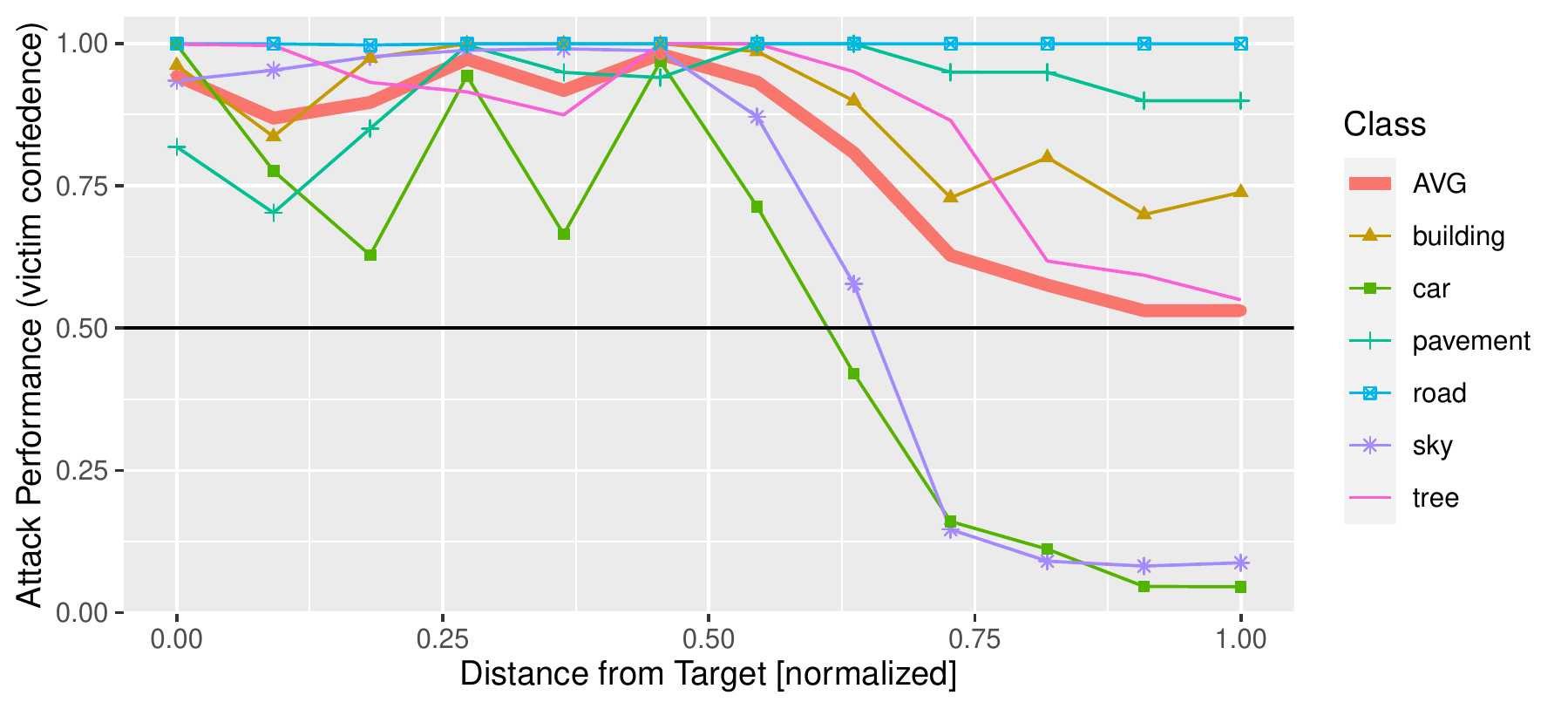}
	\caption{The average attack performance with different patch locations, shifted from near-center of $x$ to the bottom-right.}
	\label{fig:exp4a_4c}
\end{figure}

In Fig. \ref{fig:exp4a_4c} we can see that the attack is highly effective for all classes up to about 62\% of the distance away from the target (image center). The sharp drop in attack performance for the tree and sky classes is understandable since there are fewer contextual semantics which can be exploited by the patch in the bottom right of the image. On the other hand, in areas just below the horizon (0-0.5 on the x-axis), the patch can exploit contextual semantics which the model uses (e.g., features such as lighting, reflections, and building geometry). 

These results indicate that an attacker may be able to increase the likelihood of success by placing the patch on objects which have some contextual influence on the target region. For example, to create a crosswalk, it may be advantageous to put the sticker on a lamp post or parking meter since these objects may be found near crosswalks.

\subsubsection{Impact of the Target Size}
\begin{figure}[t]
	\centering
	\includegraphics[width=\columnwidth]{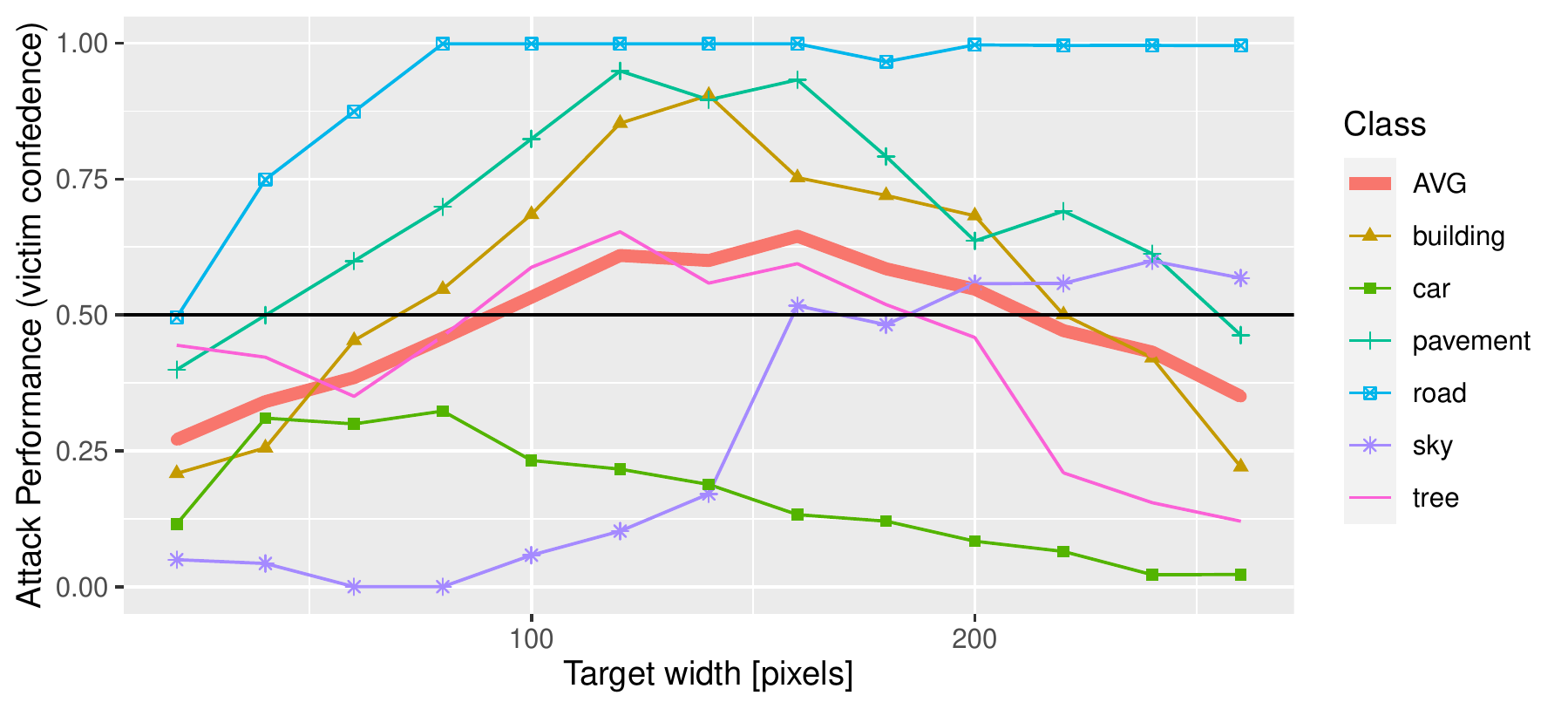}
	\caption{The average attack performance on the same 20x20 region, but with different target sizes.}
	\label{fig:exp3a_3c}
\end{figure}

In this experiment, we increased the size of $t$ but observe the performance of the same 20x20 pixel region at the center of $t$ (i.e., our objective). In Fig. \ref{fig:exp3a_3c} we can see that large targets do not perform well. The reason for this is that having a large target requires the IPatch to subdue more semantics. As a result, the patch fails and the region of $t$ becomes patchy and an corrupted. Small targets fail because it is hard for the patch to make high precision results. Rather, there is a balance between the intended 20x20 target and the actual target painted in $t$. We found that increasing the target size by a factor of 3 improves the performance at the intended region. 

The reason why a larger target helps the patch reach the 20x20 region is that the patch tends to `leech' nearby semantic regions. This makes sense since it is easier to change the boundaries of existing semantics (e.g., perceive a larger car) than generate new ones which are isolated (e.g., a tree in the middle of the road). Therefore, the added target size encourages the model to perform similar tactics.

\subsubsection{Impact of the Target Location}
In Fig. \ref{fig:exp5b} we present the attack performance when targeting different remote locations in $x$. It is clear that the influence of a patch on different regions is dependent on both the image's content and the targeted class. For example, it is easier to convince the model that any space under the horizon is a road, yet it is hard to change the class of the top-center to building because it is rarely found there. Overall, this experiment demonstrates that the patch can target locations on far remote locations within the image. However, this capability is not uniform across the classes, as we can see with the class `car'.

\begin{figure*}[t]
	\centering
	\includegraphics[width=\textwidth]{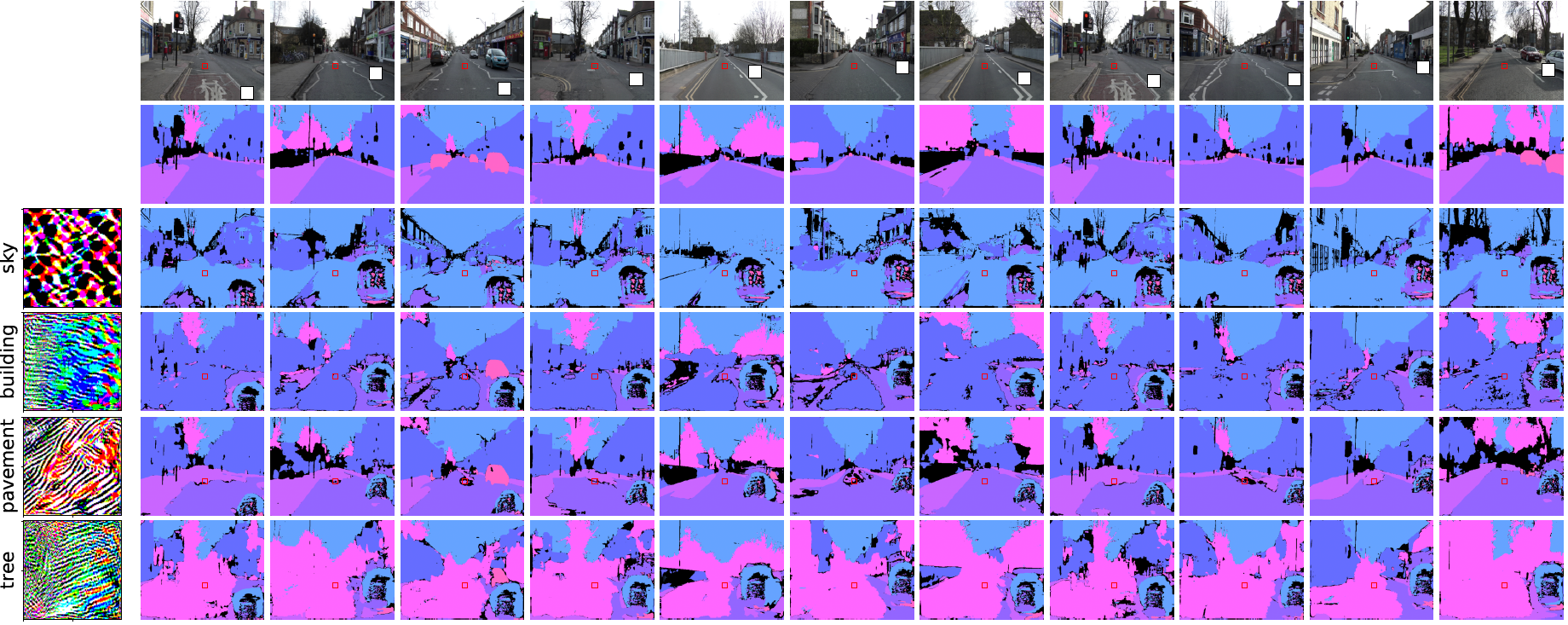}
	\includegraphics[width=0.4\textwidth]{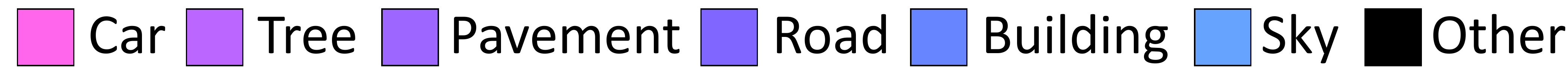}
	\caption{Examples showing the semantic segmentation maps of \texttt{efficientnet-b7\_FPN} when attacked with `robust' patches trained using EoT. Top two rows- the original image and the original segmentation map, where the target and patch are marked in red and white accordingly. Right- the IPatch used in the attack, each one has been trained for a different class but the same remote target location.}
	\label{fig:seg_grid}
\end{figure*}

\begin{figure}[t]
	\centering
	\includegraphics[width=\columnwidth]{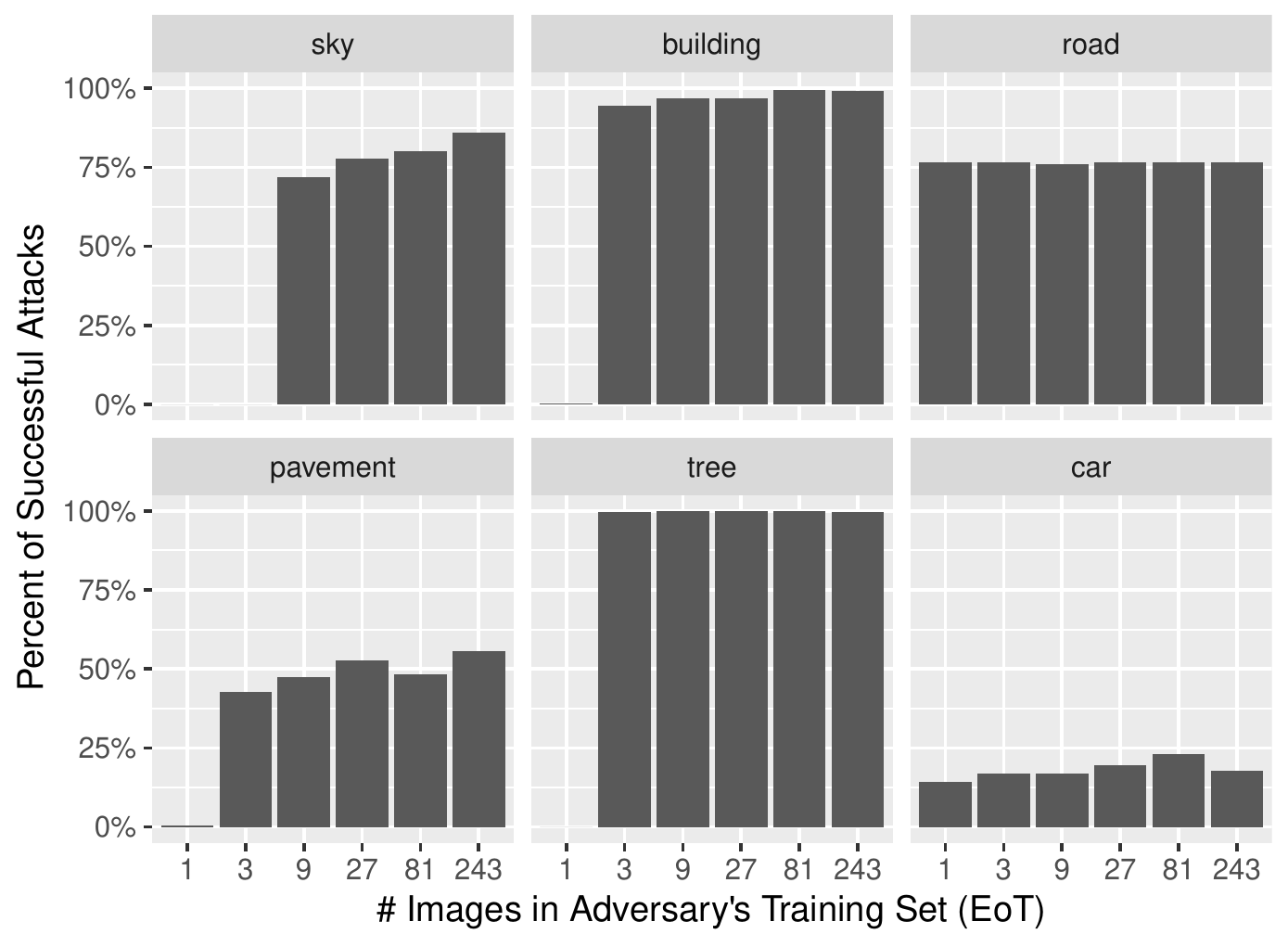}
	\caption{The performance of a robust patch when trained on different set sizes.}
	\label{fig:exp6a}
\end{figure}

\begin{figure}
	\centering
	\includegraphics[width=\columnwidth]{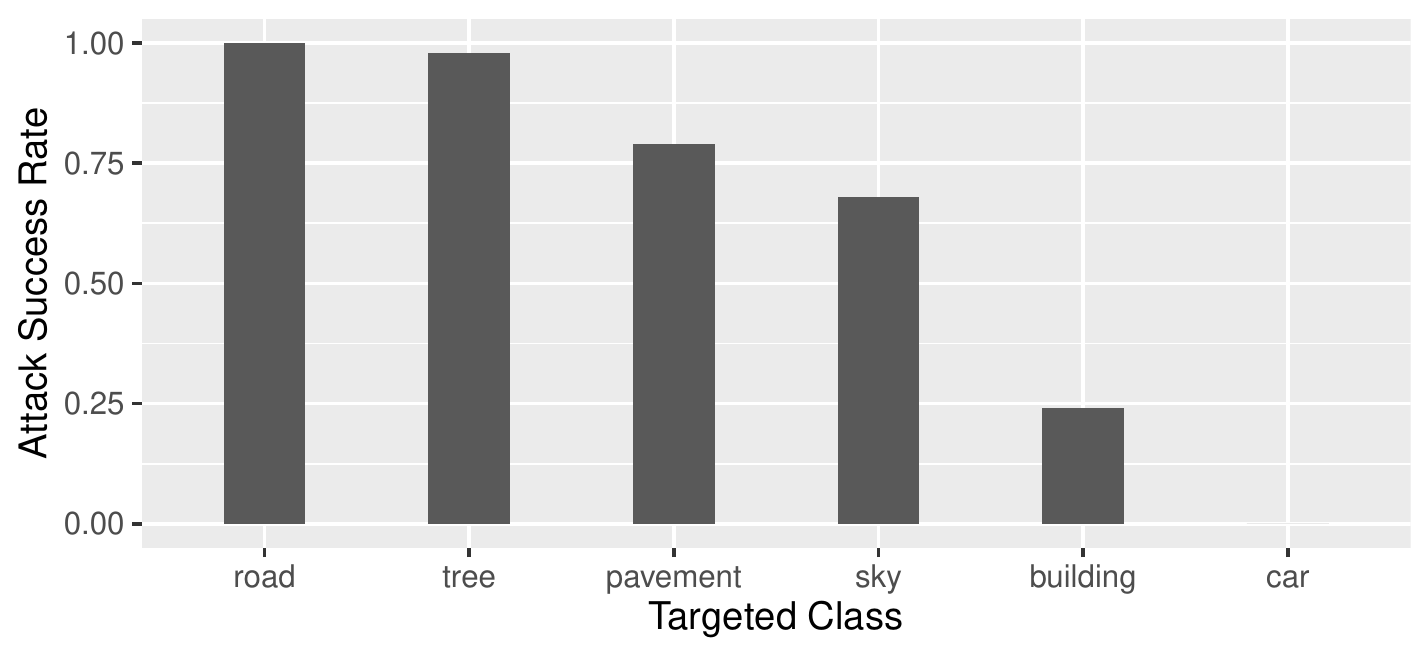}
	\caption{The average attack performance of the patches when applied to new images.}
	\label{fig:exp9a}
\end{figure}

\subsection{EXP2: Patch Robustness}\label{subsec:exp2}
In is experiment we evaluate how well a single patch performs on (1) different transformations and (2) on multiple seen and unseen images. 

\noindent\textbf{Experiment Setup.} To perform this experiment, we used EoT (\ref{eq:objective}) to train a single patch for each class. The incremental training framework from \ref{subsec:approach} was used with the patch origin $o$ set to (370,270). For the patch size $S$, we sampled uniformly on the range of [50,80] pixels. For the shifts $\ell$, we sampled uniformly within the entire bottom-right quadrant of $x$. We found that training the patch in one region helps it converge using the incremental strategy, while still generalizing to the opposite side. We targeted the same segmentation model used in EXP1, and the training was performed using a batch size of 20 (the maximum for a 24GB GPU) with a learning rate of 0.5.

\noindent\textbf{Generalization to Multiple Images}
In Fig. \ref{fig:seg_grid} we present the performance of the patches in the form of the model's perception. The images demonstrate that the IPatch generalizes well to multiple images, even at different locations and scales. In Fig. \ref{fig:exp6a} we present the attack performance when training on different numbers of images from $D_{test}$ (evaluated against the same set). From here we can see that an exponential number of examples are needed to increase the performance.

\noindent\textbf{Generalization to New Images.}
To use the patch in a real world setting, it must work well in scenes which were not in the attacker's training set. Figure \ref{fig:exp9a} presents the attack performance of patches trained on $D_{test}$ (those displayed in Fig. \ref{fig:seg_grid}) when applied to images in $D_{val}$. The results show that the patches generalize well to unseen images. More interestingly, the performance of some classes are dramatically different compared to patches trained on single images without EoT (EXP1, Fig. \ref{fig:exp1a_1b}). For example, `tree' now has a 0.98\% success rate compared to 50\% and `building' is now 25\% compared to 65\%. We learn from this that by considering multiple images, the model can learn stronger tactics. At the same time, the variability of the transformations prevent the model from using highly specific adversarial patterns. We also note that the class 'car' does not transfer to unseen images like the other classes. We attribute this to the segmentation model's poor performance on detecting cars in general.\footnote{Although the selected \texttt{efficientnet-b7\_FPN} achieves a lesser intersection over union score of 0.75 on that class, it outperforms the other models overall.}

\subsection{EXP3: The Impact on Different Models}
In is experiment we explore the suceptibilty and transferability of patches between models.

\begin{figure}[t]
	\centering
	\includegraphics[width=\columnwidth]{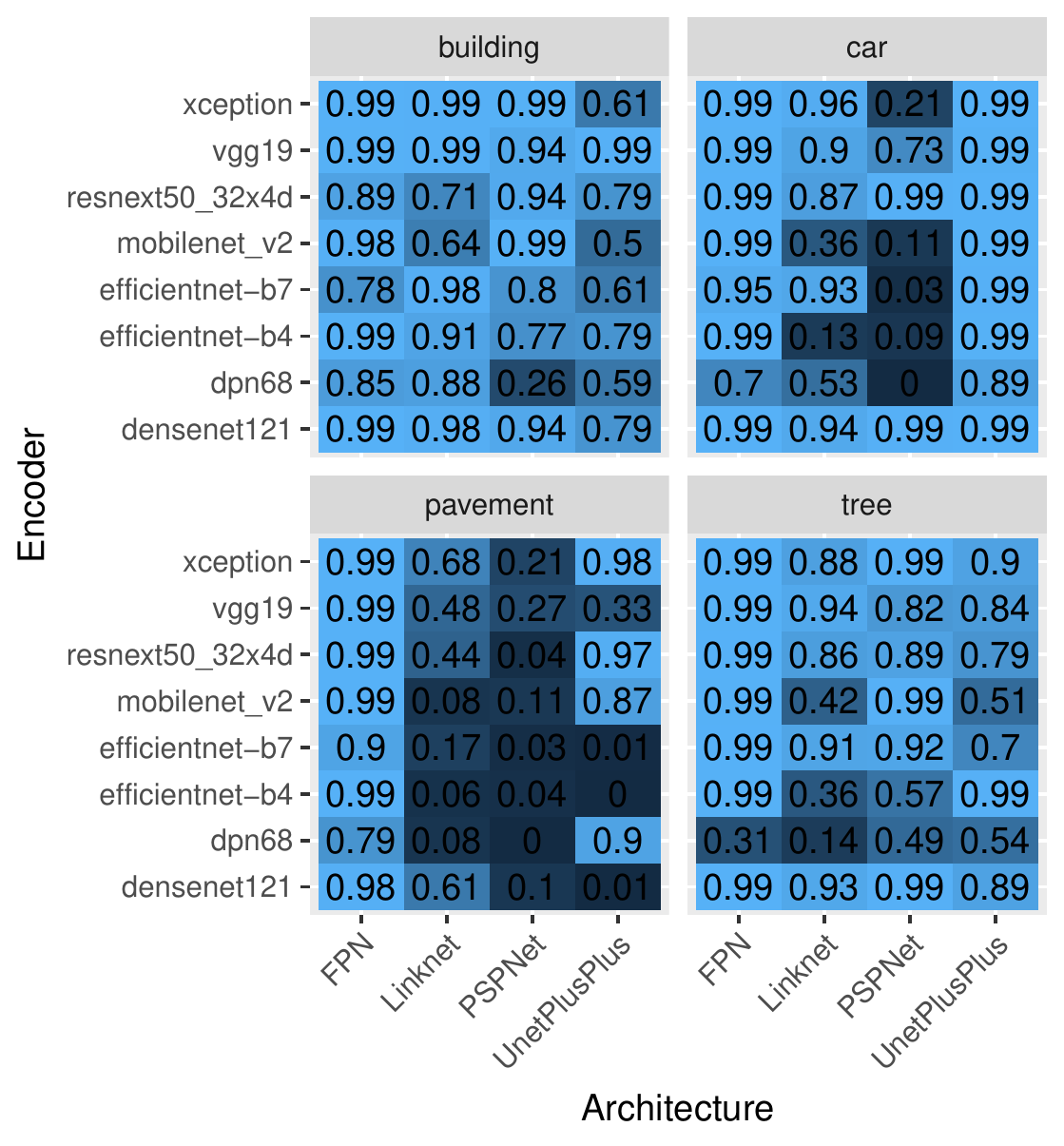}
	\caption{The average confidence of different models when attacked with an IPatch. Values greater than 0.5 indicate a successful attack on average.}
	\label{fig:exp7a}
\end{figure}
\begin{figure}[t]
	\centering
	\includegraphics[width=\columnwidth]{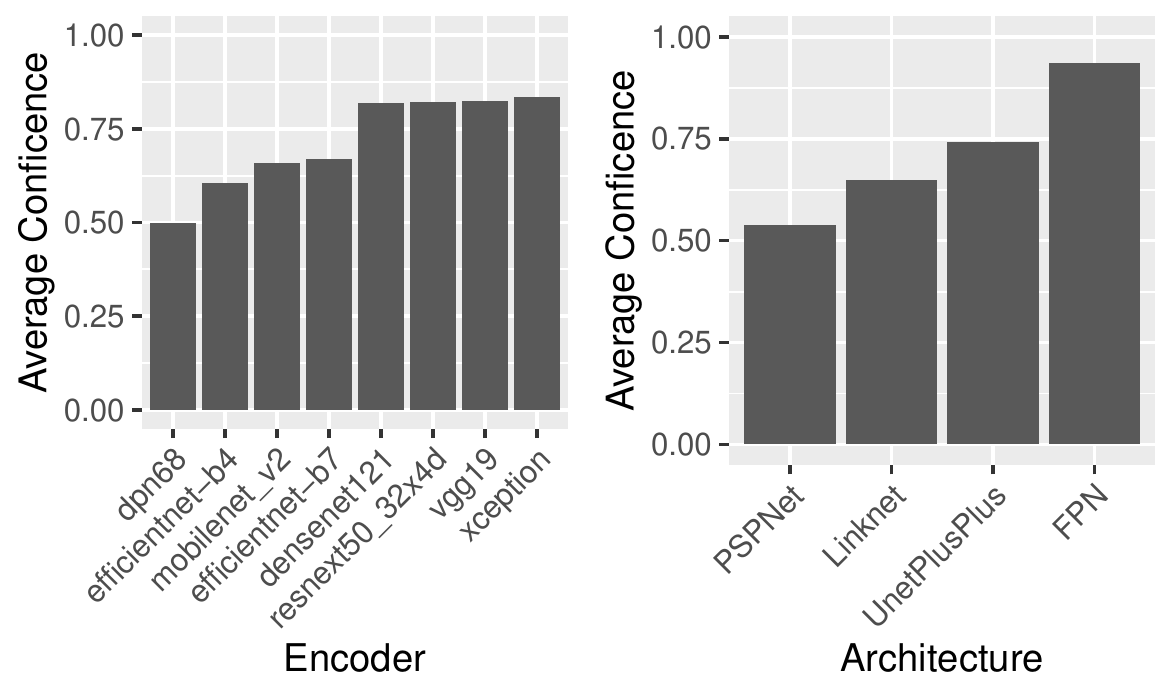}
	\caption{The average confidence of different encoders and architectures when under attack.}
	\label{fig:exp7c}
\end{figure}

\begin{figure}[t]
	\centering
	\includegraphics[width=\columnwidth]{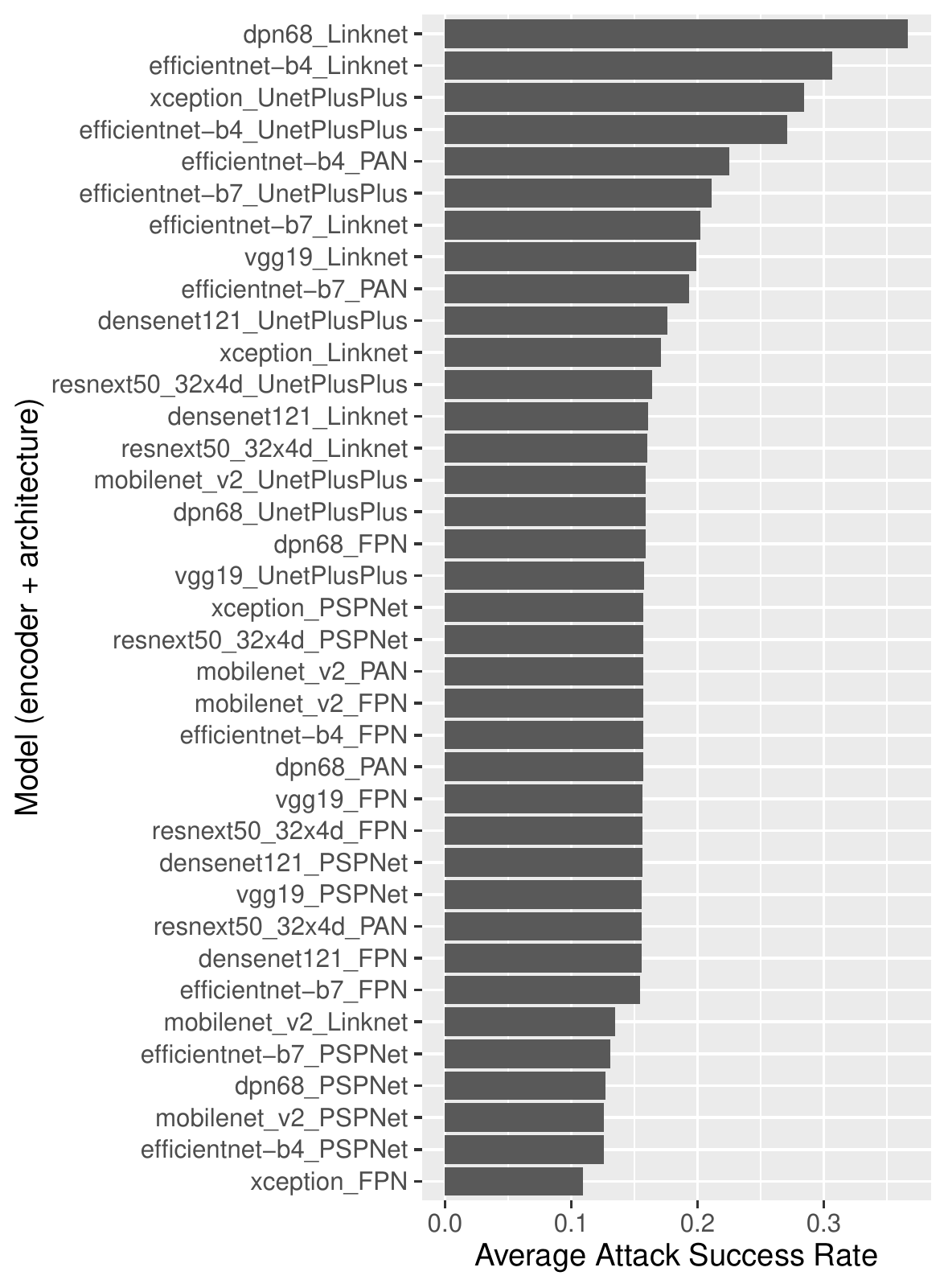}
	\caption{The average attack success rate on each model when attacked with a patch trained on \texttt{efficientnet-b7\_FPN}.}
	\label{fig:transfer}
\end{figure}

\subsubsection{Model Susceptibility}

\noindent\textbf{Experiment Setup.} To evaluate the performance of the attack on different model architectures, we used 32 of the 37 segmentation models described at the beginning of section \ref{sec:eval} (\texttt{PAN} was omitted since it was not compatible with Torch's autograd in our framework). Due to time limitations, the attacks on each model were limited to 4 classes, 10 images, and 3 minutes training time for each image. 

\noindent\textbf{Results.}
We found that all 36 models are susceptible to the RAP attack for at least one class(Fig. \ref{fig:exp7a}). By observing the patterns in the columns, we note that some architectures are less susceptible to attacks on certain classes. For example, \texttt{Linknet}, \texttt{PSPNet}, and \texttt{Unet++} on pavement and \texttt{PSPNet} on car. 

In Fig. \ref{fig:exp7c}, we can see the suceptibilty of the encoders and architectures overall. Some of the most susceptible encoders (\texttt{xception} and \texttt{resnext}) and architectures (\texttt{FPN} and \texttt{Unet++}) use skip connections or residual pathways in their networks. These pathways enable the networks to capture features at multiple scales and capture the global contexts better. However, just as these network utilize these pathways to obtain better perspectives, so can the IPatch in order to reach deeper into the image. Interestingly we found that the \texttt{dpn68} encoder is consistently resilient against the attack. This encoder is formally called a Dual Path Network \cite{chen2017dual}. It uses a residual path like a \texttt{ResNet} to reuse learned features and a densely connected path like \texttt{DenseNet} to encourage the network to explore new features. These diverse features may be preventing the IPatch, and possibly the segmentation model, from reaching remote contexts.

\subsubsection{Inter-Model Transferability}
In the case where the attacker does not have knowledge of the victim's model, we would like to know well a patch trained on one model transfers to others. 

\noindent\textbf{Experiment Setup.} To perform this experiment, we took the robust patches trained using the \texttt{efficientnet-b7\_FPN} (EXP2) and attacked each of the other 36 models (listed in Fig. \ref{fig:transfer}). The patches for the top 4 classes (sky, building, pavement, tree) were applied to the images in $D_{test}$ using the random transformations described in EXP2.

\noindent\textbf{Results.} We found that patch from \texttt{efficientnet-b7\_FPN} can influence the other models' predictions on the target region with an attack success rate of 11-37\% (about 1-4 times in every 10 cases). We note that a cameras on an autonomous car processes at least 30 frames per second. Therefore, there is a high likelihood that the car's model will be susceptible to the attack while driving by. 

Fig. \ref{fig:transfer} shows the largest confidences for each model, measured as the relative increase from the original confidence (on clean image). Interestingly models using the \texttt{Unet++} architecture were the most susceptible, followed by \texttt{Linknet}. We believe the reason for this is that both of these models use skip-connections to allow for feature maps to bypass the encoding process. As a result, features in the patch have a more direct impact on the output. It is known that skip-connections make models more vulnerable to adversarial examples \cite{wu2020skip} but it is interesting to see that they are vulnerable to transfer attacks as well. Another observation is that there does not seem to be a correlation between the results and the encoder used. This is probably because all of the encoders were trained on the same ImageNet Dataset.

Finally, we note that there are ways in which an adversarial example can be made to be more transferable \cite{xie2019improving,huang2019enhancing}. We leave this for future work.

\section{Extending to Object Recognition}
In section \ref{sec:eval} we performed an in-depth evaluation and analysis of the IPatch as a RAP against segmentation models. However, the same training framework in \ref{subsec:approach} can be used on other semantic models as well. In this section, we present preliminary results against a popular object recognition model called \texttt{YOLOv3} \cite{redmon2018yolov3}.

\subsection{Technical Background}
\begin{figure}[t]
	\centering
	\includegraphics[width=\columnwidth]{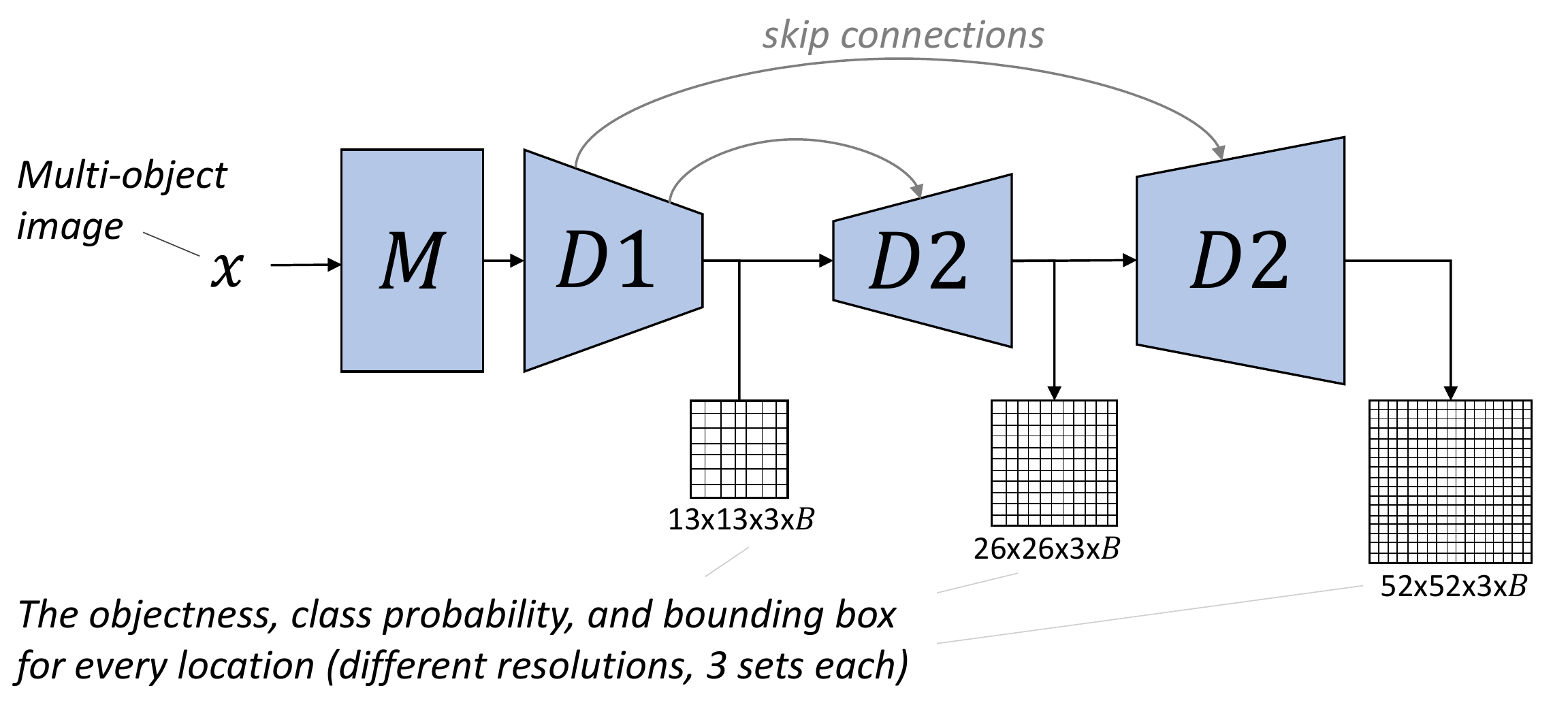}
	\caption{The architecture of the YOLOv3 object detector.}
	\label{fig:yolo}
\end{figure}

The family of YOLO models follow a similar architecture (Fig. \ref{fig:yolo}). $x$ is passed through a series of convolutional layers ($M$ in the figure) and then those feature maps are shuttled to various decoders. The decoders predict coarse maps to the image at different scales using the semantic information shared between them. The multiple scales help the model detect objects of different sizes (e.g., $D1$ detects large objects). Each cell in a map, contains an objectness score, class probability, and a bounding box (obtained via regression). If a cell has an objectness score above some threshold, then there is an object there with the associated class probability. Finally, a non-maximal suppression (NMS) algorithm is used on the maps to identify and unify the detections. 

\subsection{Evaluation}
\noindent\textbf{Experiment Setup.} To see if the attack would work on YOLO, we created an IPatch which convinces YOLO that there is a person standing in the middle of the road. To accomplish this, we used a pre-trained \texttt{YOLOv3} model implementation\footnote{\url{https://github.com/eriklindernoren/PyTorch-YOLOv3}} as the victim, and trained our patch using 30k images: 15k random samples from the Bdd100k dataset \cite{yu2020bdd100k} and 15k frames from a Toronto car driving video on YouTube\footnote{\url{https://youtu.be/50Uf_T12OGY}}. 

For training, we needed to ensure that both the objectness score and probability of the class `person' were high. This was done by taking the product of $D1$'s probability map and objectness map as $y_s'$, and by setting $t$ to highlight the cell in the lower-center of the image. For the loss functions, we took the sum of $\mathcal{L}_{KL}$ and $\mathcal{L}_{1}$ since it increased the rate of convergence. EoT was used to scale the patch between 60-70 pixels in width and shift it randomly within the bottom-right quadrant of the image. Finally, we trained the patch for 3 days with a learning rate of 0.05. 

\begin{figure}[t]
	\centering
	\includegraphics[width=\columnwidth]{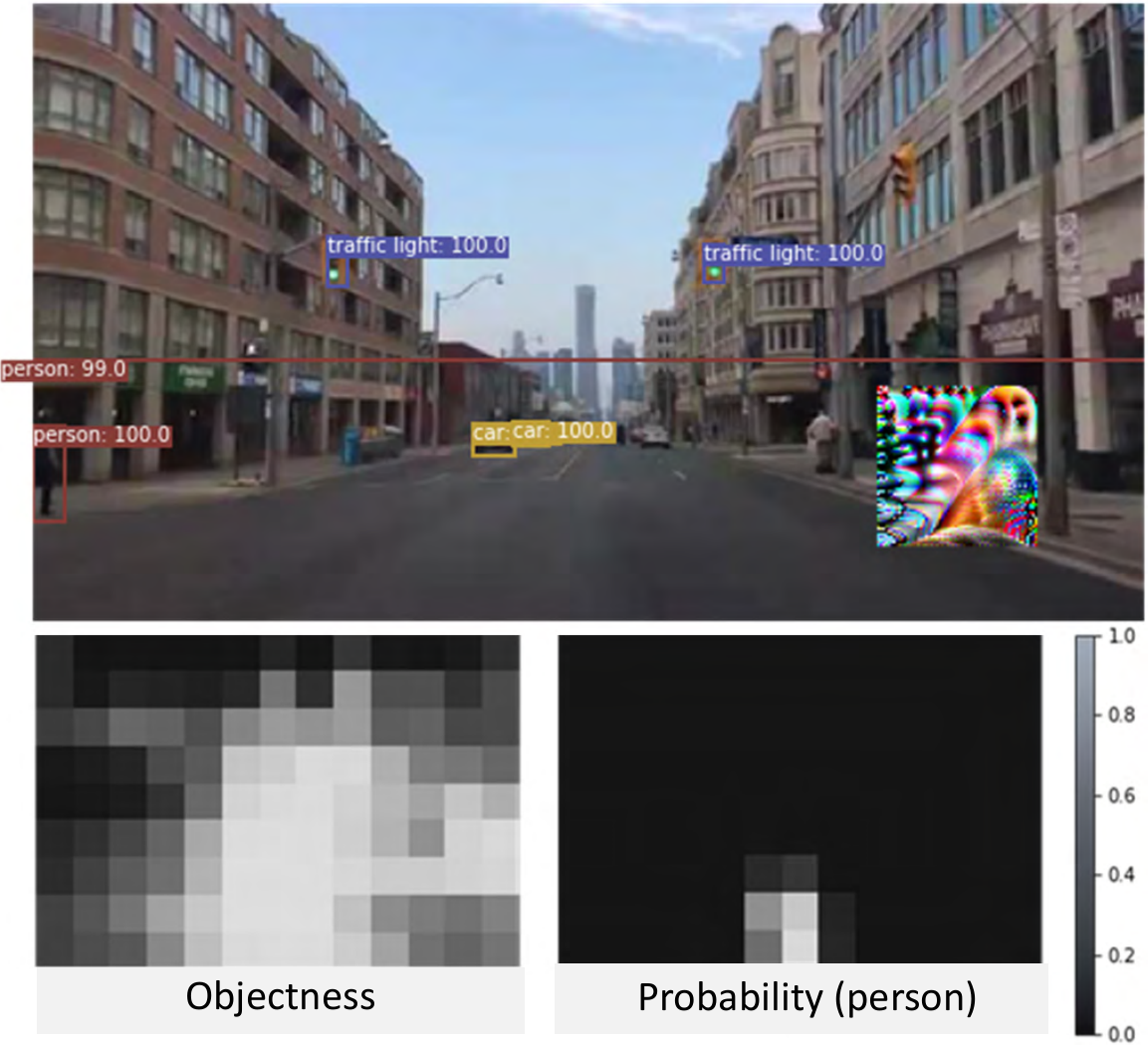}
	\caption{An example of the \texttt{YOLOv3} object detector being attacked by an IPatch. Here, the patch has convinced the model that there is a person standing in the middle of the road.}
	\label{fig:yolo_result}
\end{figure}

\noindent\textbf{Results.}
We found that the \texttt{YOLOv3} object detector is susceptible to the attack with an 85\% attack success rate. In Fig. \ref{fig:yolo_result} we present an example frame which shows the objectness and probability maps in $D1$ during the attack. We also found that smaller patches ranging from 50-60 pixels in width achieve an 80\% attack success rate. Overall, it was relatively easy for the framework to change the objectness score of arbitrary locations in the image, compared to the class probability. We also observed that it is significantly harder to target the maps from $D2$ and $D3$ which capture smaller objects. We believe this is because $D2$ and $D3$ rely less on contexts around the image, giving the IPatch less leverage to perform a remote attack. 

As future work, we plan to explore RAPs on other object detectors and investigate other semantic models as well.

\section{Discussion \& Countermeasures}\label{sec:dicussion}

The concept of a remote adversarial patch, introduced in this paper, opens up wide range of possible attack vectors against image-based semantic models. Through our observations in \ref{sec:eval}, we were able to identify some of the attack's capabilities and limitations. 

\noindent\textbf{Trade-Offs.} Due to its flexibility, it may seem like the IPatch is harder to defend against compared to an ordinary adversarial patch. However, the performance of the patch is less compared to a 'point-based' patch. This means that the adversary must consider whether a more reliable attack needed over having flexibility and stealth. Another consideration is that the adversary may want to experiment to find the optimal placement of the patch. This is because some regions give the patch more leverage based on the local semantics (section \ref{subsec:patch_location}). One strategy is that the attacker can first scout the target region by videoing the scene from multiple perspectives and then optimize the patch location using that dataset.

\noindent\textbf{Defenses.} Although the IPatch can be placed in arbitrary locations, we noticed that its presence highly noticeable in the semantic segmentations (e.g., Fig. \ref{fig:seg_grid}). We found that it is very hard to generate a patch which both achieves the attack and masks its own presence at the same time. Concretely, when setting $t=y_s'$ except for the target region (as done in Fig \ref{fig:arbitrary}), we found that the model struggles to influence remote locations to the same extent. In future work, this may be improved through a custom loss function which balances the trade-off between the two objectives. Another solution might be to generate RAPs using a conditional GAN which considers the errors on the semantic map in ($\neg m$). Doing so may also reduce the corruptions to nearby semantics as well. 

Another direction for defending against this attack is to limit the model's dependency on global features. Although these global features are key to state-of-the-art models \cite{lin2017feature,li2018pyramid,9201310}, it is possible to utilize them while also considering their layout and origin. One option may be to integrate capsule networks \cite{sabour2017dynamic} as part of the model's architecture, since capsule networks are good at considering the spatial relationship in images. 

\noindent\textbf{Improvements.}
We noticed that the RAP attack is dependent on an image's content when targeting segmentation models, but less so for the object detector YOLO. For example, we were able to perform remote adversarial attacks on a blank image $x$ with YOLO. The reason for this is not clear to us, and investigating it may lead to improvements in the proposed training methodology. Moreover, as future work, it would be interesting to investigate which types of features and classes the a RAP can manipulate best and why. This research may lead to deeper insights into the vulnerability's extents and limitations. Finally, to improve transferabilty, we suggest two directions: (1) include multiple models in the training loop to help the model identify common features, and (2) use adversarial training to improve the generalization of the patch. 

We also noticed that performance improves when the pixel values of the patch are not clipped to the range 0-255. This is because larger inputs enable the patch to exploit  weaker contextual pathways by saturating activations in the network. As future work, it would be interesting to see if an incremental strategy can be used on the clipping range to ultimately produce stronger patches.

\section{Conclusion}

In this paper, we have introduced the concept of a `remote adversarial patch' (RAP) which can alter the semantic interpretation of an image while being placed anywhere within the field of view. We have implemented an RAP called IPatch and demonstrated that it is robust, can generalize to new scenes, and can impact other semantic models such as object detectors. With an average attack success rate of up to 93\%, this attack forms a tangible threat. Although RAPs are in their infancy, we hope that this paper has laid some of the groundwork for exploring this new adversarial example.

In summary, neural networks are notorious for being black-boxes which are difficult to interpret. However, they are still used in critical tasks because their advantages outweigh their potential disadvantages. We hope that our findings will help the community improve the security of deep learning applications so that we may continue to benefit from safe and reliable autonomous systems. 

\bibliographystyle{IEEEtran}
\bibliography{paper}

\begin{IEEEbiographynophoto}{Yisroel Mirsky}
	is a tenure-track lecturer in the Department of Software and Information Systems Engineering at Ben-Gurion University. He received his Ph.D. from BGU in 2018 and was a postdoctoral fellow for two years in the at the Georgia Institute of Technology in the research labs of Prof. Wenke Lee. His main research interests include deepfakes, adversarial machine learning, anomaly detection, and intrusion detection. Dr. Mirsky has published his work in some of the best security venues: USENIX, CCS, NDSS, Euro S\&P, Black Hat, DEF CON, RSA, CSF, AISec, etc. His research has also been featured in many well-known media outlets: Popular Science, Scientific American, Wired, The Wall Street Journal, Forbes, and BBC. Some of his works, include the exposure of vulnerabilities in the US 911 emergency services and research into the threat of deepfakes in medical scans, both featured in The Washington Post.
\end{IEEEbiographynophoto}

\vfill

%
%
%
%
%
%
%
%

\end{document}